\documentclass{article}
\usepackage{microtype,graphicx,subfigure,booktabs,hyperref}

\usepackage[accepted]{icml2024}

\usepackage{amsmath,amssymb,amsthm,mathtools,txfonts}
\usepackage[capitalize,noabbrev]{cleveref}

\usepackage{enumitem}
\usepackage{makecell}

\usepackage{hyperref}
\hypersetup{
    colorlinks=true,
    linkcolor=blue,
    filecolor=magenta,      
    urlcolor=cyan,
    pdftitle={Overleaf Example},
    pdfpagemode=FullScreen,
    }

\theoremstyle{plain}

\theoremstyle{definition}

\theoremstyle{remark}

\usepackage[textsize=tiny]{todonotes}
\usepackage{tcolorbox,xcolor}
\definecolor{c_ready}{rgb}{0.0, 0.0, 1.0}

\definecolor{purp}{rgb}{0.65, 0.16, 0.65}

\newcommand{\eg}{\textit{e.g.}}
\newcommand{\ie}{\textit{i.e.}}

\icmltitlerunning{Neural Image Compression with Text-guided Encoding}

\begin{document}

\twocolumn[
\icmltitle{Neural Image Compression with Text-guided Encoding\\for both Pixel-level and Perceptual Fidelity}
%\icmltitle{TACO: Text-Adaptive Encoding for Neural Image Compression}

\icmlsetsymbol{equal}{*}

\begin{icmlauthorlist}
\icmlauthor{Hagyeong Lee}{equal,postech}
\icmlauthor{Minkyu Kim}{equal,postech}
\icmlauthor{Jun-Hyuk Kim}{sait}
\icmlauthor{Seungeon Kim}{sait}
\icmlauthor{Dokwan Oh}{sait}
\icmlauthor{Jaeho Lee}{postech}
\end{icmlauthorlist}
\icmlaffiliation{postech}{POSTECH}
\icmlaffiliation{sait}{Samsung Advanced Institute of Technology}

\icmlcorrespondingauthor{Jaeho Lee}{jaeho.lee@postech.ac.kr}
\icmlkeywords{Image compression, Multi-modal learning}

\vskip 0.3in
]

\printAffiliationsAndNotice{\icmlEqualContribution}

\begin{abstract}
Recent advances in text-guided image compression have shown great potential to enhance the perceptual quality of reconstructed images. These methods, however, tend to have significantly degraded pixel-wise fidelity, limiting their practicality.
To fill this gap, we develop a new text-guided image compression algorithm that achieves both high perceptual and pixel-wise fidelity.
In particular, we propose a compression framework that leverages text information mainly by text-adaptive encoding and training with joint image-text loss. By doing so, we avoid decoding based on text-guided generative models---known for high generative diversity---and effectively utilize the semantic information of text at a global level. Experimental results on various datasets show that our method can achieve high pixel-level and perceptual quality, with either human- or machine-generated captions. In particular, our method outperforms all baselines in terms of LPIPS, with some room for even more improvements when we use more carefully generated captions.

% \footnote{Our code is available at: \href{https://taco-nic.github.io}{taco-nic.github.io}}

% \href{https://taco-nic.github.io}{taco-nic.github.io}

\begingroup%
\fontsize{8.5pt}{10pt}\selectfont%
\noindent\begin{tabular}{@{}lr@{}}
\textbf{\fontsize{9.5pt}{11pt}\selectfont Project Page:}\hspace{3.3em} & \hspace{-1.025em} 
\href{https://taco-nic.github.io}
{\texttt{taco-nic.github.io}}\\[0.8ex]

{\textbf{\fontsize{9.5pt}{11pt}\selectfont Code:}} & \hspace{-1.025em}%
{
\href{https://github.com/effl-lab/TACO}%
{\texttt{github.com/effl-lab/TACO}}%
}
\end{tabular}%
\endgroup%

\end{abstract}
\label{submission}

\section{Introduction}\label{sec:intro}

\begin{figure}[!t]
    \centerline{\includegraphics[width=\columnwidth]{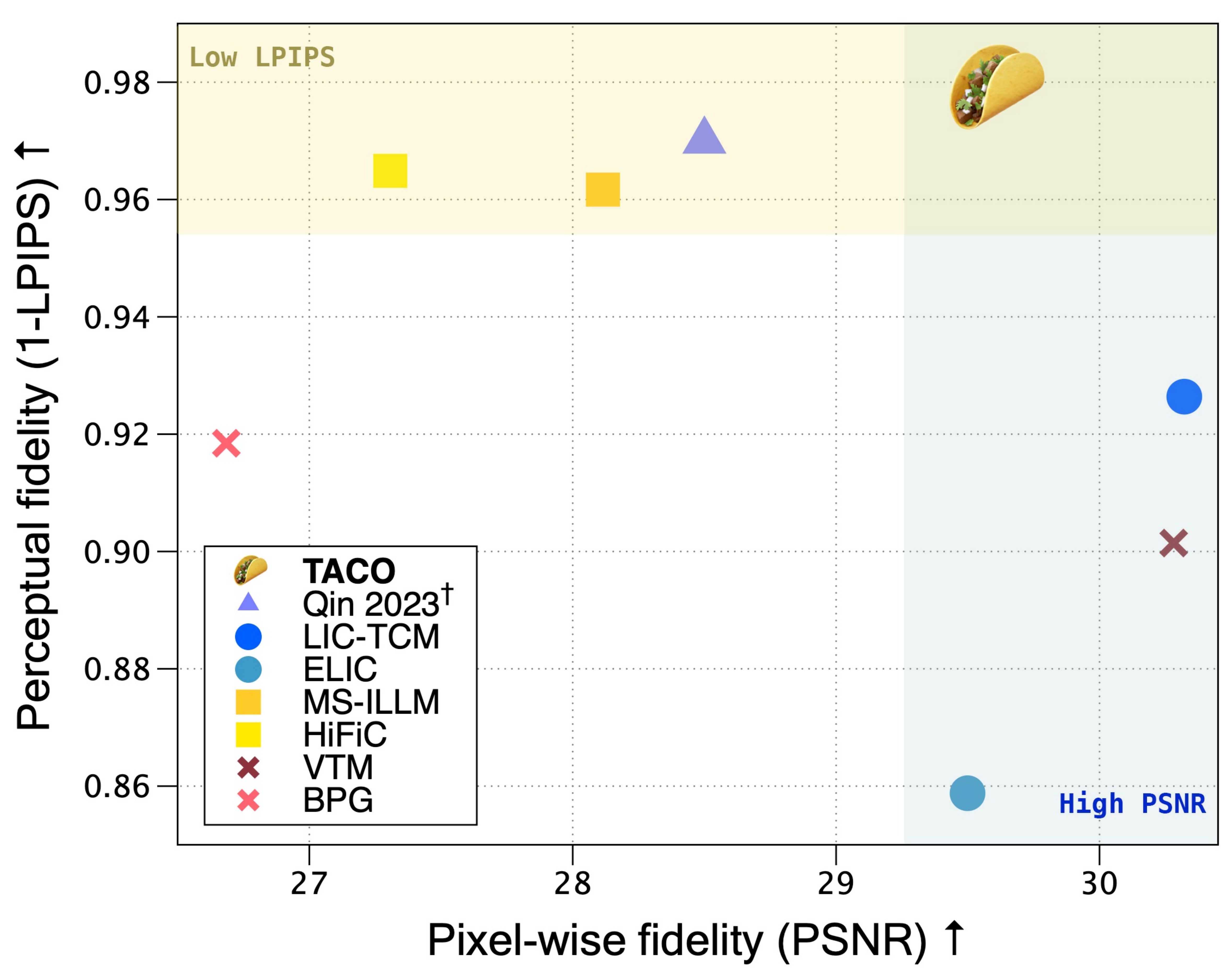}} % 
    \caption{\textbf{Pixel-wise fidelity \textit{vs.} perceptual fidelity, at 0.40 bpp.} We compare pixel-wise and perceptual fidelity of image compression codecs on MS-COCO 30k. The proposed TACO achieves competitive results in both metrics. The reported figures for Qin 2023 has been measured on MS-COCO 40k, and some figures have been interpolated from nearest bpp models 
    ($\medbullet$: PSNR-focused, $\blacksquare$: perception-focused, $\blacktriangle$: text-guided, $\times$: handcrafted).}
    \label{fig:fig1}
\end{figure}

% , and the figures of PerCo has been extrapolated from lower-bpp results 

How can text improve machine vision? This question has been asked repeatedly in various domains of visual computing, giving birth to a number of vision-language models with tremendous multi-modal reasoning and generative capabilities \citep{clip,llava}.

The value of text has gained much attention in \textit{image compression} as well, motivated by the astonishing ability of modern text-guided generative models to synthesize realistic images from the given text prompt \citep{latent_diffusion}. The conventional way to leverage these models is via text-guided decoding (or \textit{textual transform coding}; \citet{textual}): We use text-guided generative model as our decoder, and the encoding is done by finding a good textual code. The code may consist of a text prompt plus any additional latent prompt vectors, that can generate the reconstruction that looks close to the original image. As the text can be represented with a very small number of bits, we expect such an approach to be very bit-efficient \citep{shannon51,bhown18,textual}.

\begin{figure*}[!t]
    \centerline{\includegraphics[width=1.86\columnwidth]{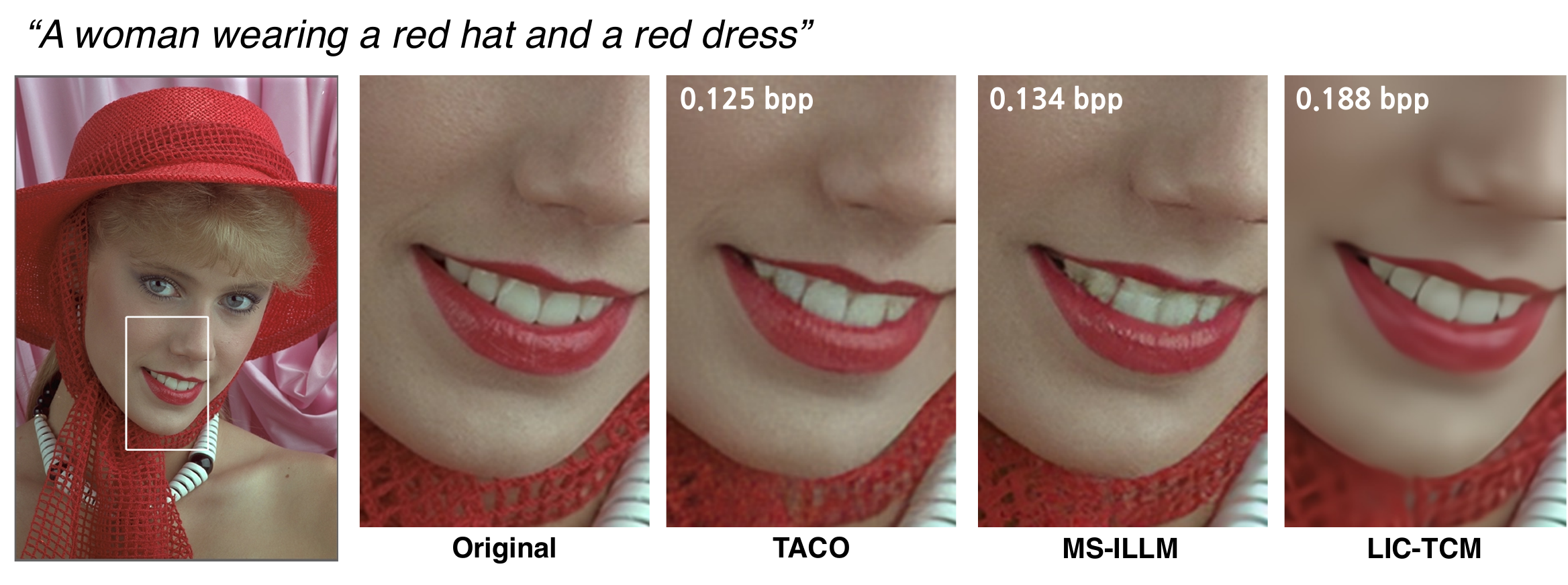}}
\caption{\textbf{Qualitative results.} We compare TACO against MS-ILLM and LIC-TCM, which focuses on perception and PSNR, respectively. TACO uses slightly less bpp than baselines. Comparing with MS-ILLM, TACO tends to suffer less from hallucinated artifacts (see teeth). Comparing with LIC-TCM, TACO can reconstruct sharper details (see lips). See \cref{app:add_qualitative_examples} for more examples.}\label{fig:qualitative}
\vspace{-0.2em}
\end{figure*}

Following this intuition, recent works have developed text-guided neural image compression codecs that can reconstruct images with high \textit{perceptual quality}, \ie, either highly realistic (\eg, low FID) or perceptually similar to the original image (\eg, low LPIPS). One line of work shows that using pretrained text-guided generative models as decoders can give neural codecs that achieve high realism at extremely low bitrates, \eg, less than 0.01 bits per pixel \citep{pan_diffusion_22,textsketch,careil24}. Another line of work targets conventional bitrates (over 0.1 bpp) and shows that one can significantly improve the perceptual similarity of reconstructed images to the original image by using text captions as additional side information, where the text information is inserted to the decoder via some plug-in modules \citep{tgic,qin_comp_23}.

These approaches, however, commonly suffer from substantial degradations in pixel-wise fidelity, \ie, PSNR. For example, diffusion-based compression codecs tend to achieve 3–5 dB less PSNR than standard neural image compression codecs \citep{careil24}. Despite the practical importance of the pixel-wise fidelity \citep{lictcm}, it remains unknown whether or how such PSNR degradation can be mitigated for text-guided image compression schemes.

\textbf{Contribution.} In this paper, we fill this gap by developing a text-guided image compression scheme that can achieve high pixel-level and perceptual quality simultaneously (we focus more on the perceptual fidelity, \eg, LPIPS, than on realism). Our key hypothesis is that text-guided decoding \textit{may not} be an effective strategy for PSNR. For high PSNR, we must be able to decode with very small generative diversity, and effectively utilize the semantic information at the global level (\ie, not affecting only local regions). Diffusion-like decoders tend to have too much diversity \citep{pan_diffusion_22}, and plug-in approaches tend to have weak control on the global semantics of the reconstruction \citep{textsketch}.

Thus, we establish an alternative strategy of utilizing text \textit{\textbf{only for encoding}}. Here, the main role of text is to provide additional supervision on \textit{how human perceives the image}; using this information, the encoder can better preserve perceptually meaningful information without discarding it during (lossy) compression. By using text only for encoder, we can leverage existing decoder architectures that can generate accurate reconstructions with minimal generative diversity. Furthermore, as we encode the textual information into the code itself, the injected semantic information can affect the whole pixels globally without any extra effort.

Based on this intuition, we propose a simple yet effective encoder-centric text-guided image compression algorithm, coined \textbf{TACO} (\underline{T}ext-\underline{A}daptive \underline{CO}mpression). TACO transforms a popular PSNR-oriented neural codec architecture (ELIC) into a text-guided one by augmenting the encoder with a text adapter. The text adapter utilizes a pre-trained CLIP encoder \citep{clip} and bi-directional attention mechanism to inject textual information into the latent code. Then, TACO trains the model from scratch with a joint image-text loss, which encourages the reconstruction to be semantically aligned with the given text.

In our experiments, we find that TACO achieves excellent performance in both pixel-level and perceptual quality at standard bpp levels (\ie, $\ge$ 0.1). In particular, TACO outperforms all image compression baselines in terms of LPIPS, while achieving competitive performance in pixel-level fidelity and realism when compared to state-of-the-art methods in each metric (\cref{fig:fig1}). This trend holds true for either datasets that come with human-generated image descriptions (\eg, MS-COCO \citep{mscoco}), or image-only datasets paired with machine-generated captions (\eg, CLIC \citep{clic2020} captioned by OFA \citep{ofa}). TACO also outperforms conventional LPIPS-focused text-guided baselines \citep{tgic,qin_comp_23}.

Our findings suggest that the core value of text in image compression---at least in standard bpp range---may hinge more on its relationship with human perception, than it being an efficient way to store information \citep{textual}.

\section{Related Work}\label{sec:related}

\textbf{Measuring Reconstruction Quality.}
The most popular metric for evaluating the quality of reconstructions is the \textit{pixel-wise distortion} from the original image. Typically, the quantity is measured in mean squared error (MSE) or peak signal-to-noise ratio (PSNR). However, the pixel-wise distortion is often found ill-aligned with the human-perceived image quality \citep{eskicioglu95}.

To address this, various \textit{perceptual quality} metrics have been proposed. Roughly, the metrics fall into two categories: \textit{perceptual distortion} and \textit{realism}. Perceptual distortion quantifies how different two images are for human perception. Multi-scale structural similarity \citep[MS-SSIM]{ms_ssim} measures the discrepancy of patch-level statistics of both images in various scales. Learned perceptual image patch similarity \citep[LPIPS]{lpips} measures the distortion in the feature space of pretrained neural net classifiers, which are fine-tuned to account for human perception. Similarly, PieAPP trains a model to approximate the human judgement on perceptual similarity \citep{prashnani18}.

Realism, on the other hand, attempts to quantify how realistic an image is, without necessarily comparing it with the original image. For example, Fr\'{e}chet inception distance \citep[FID]{fid} measures the distance of (Inception) feature distribution of reconstructed images, from that of natural images. Similarly, KID measures the maximum mean discrepancy between the feature distributions \citep{kid}. A recent work by \citet{cmmd} argues that FID/KID are still misaligned with human perception, and proposes CMMD, an alternative metric that utilizes pre-trained multimodal embeddings.

\textbf{Neural Image Compression.} An early work by \citet{balle17} proposes an autoencoder-like image compression pipeline which achieves competitive pixel-level fidelity and substantially stronger perceptual quality than handcrafted codecs, \textit{e.g.}, JPEG. Subsequent works have developed more advanced hyperpriors on the latent space codes and utilized more advanced autoencoder-like architectures \citep{balle2018,mbt18,cheng2020,elic,lictcm}, which led to even smaller pixel-wise and perceptual distortions on the reproduced image.

Another line of work focuses on developing neural compression codecs for improved realism. \citet{hific} propose a GAN-based method to generate highly realistic images at low bitrates. More recent approaches use diffusion model to generate even more realistic images \citep{hoogeboom_diffusion_23, yang_diffusion_23}. However, these methods tend to have greater pixel-level or perceptual distortions than autoencoder-based models. MS-ILLM shows that the tradeoff between the distortion and realism can be more alleviated by using a more tailored GAN discriminator \citep{ms_illm}. The fundamental limit of this rate-distortion-realism tradeoff has been theoretically studied by \citet{blau2018perception}, suggesting that both goals may not be achieved simultaneously given a fixed rate.

\textbf{Text-guided Image Compression.}
Following the great success of vision-language models, text-guided image compression algorithms have emerged \citep{bhown18,textual}. One line of work aims to leverage pre-trained text-guided generative models as decoders. Here, the key is to introduce an appropriate means to control the generation diversity of these models.  \citet{pan_diffusion_22} utilize the stable diffusion \citep{latent_diffusion}, and control the diversity by guiding the reverse step with a JPEG-compressed version of the image. \citet{textsketch} uses ControlNet as the decoder \citep{zhang23}, where the additional control is done by the `sketch' version of the image, extracted by an edge detector. A concurrent work by \citet{careil24} also uses the latent diffusion model, and proposes a more sophisticated local-global encoding for better reconstruction. These models show superior performance in terms of realism (\eg, FID), especially at extremely low bpp.

\begin{figure}[t]
    \centering
    \includegraphics[width=\columnwidth]{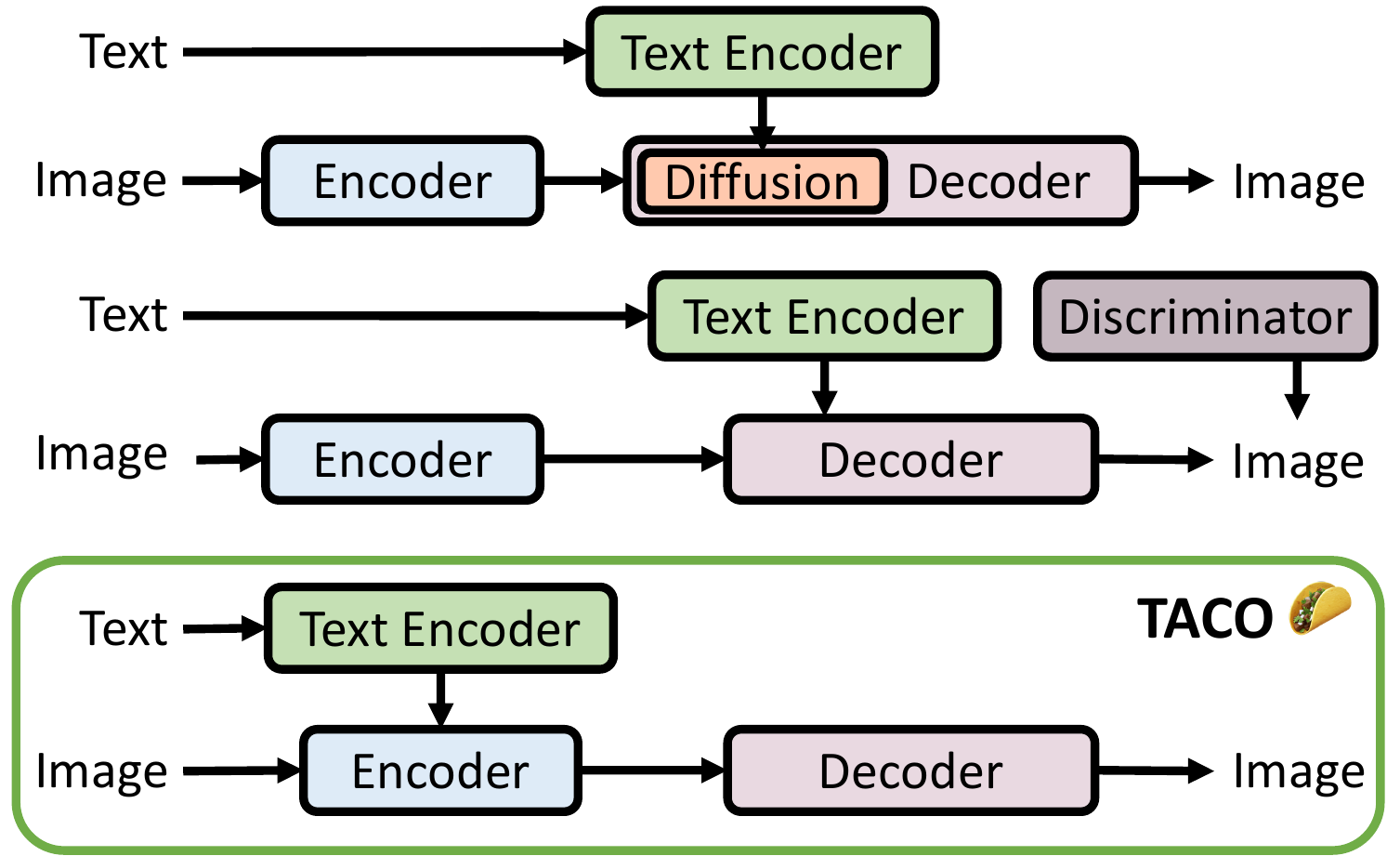}
    \caption{\textbf{Text-guided decoding strategies vs. TACO.} (Top) Text-guided decoding with diffusion-based decoders \citep{careil24}. (Middle) Text-guided decoder utilizing GAN \citep{qin_comp_23}. (Bottom) TACO is a much simpler yet effective strategy.}
    \label{fig:comparison_approaches}
\end{figure}

Another line of work proposes to use decoders that are trained from scratch, using a dataset with image-text pairs such as MS-COCO \citep{mscoco}. The network architecture typically follows GAN-based image compression codecs \citep{hific}, with additional components to insert textual information into the architectures. \citet{tgic} proposes to insert textual information to both encoder and decoder through uni-directional attention-like modules. \citet{qin_comp_23} insert text only to the decoder through semantic-spatial aware blocks. These works have their main strength on perceptual distortion (\eg, LPIPS) at conventional bpp.

In both lines of work, it has been discovered that one can achieve high realism or perceptual similarity, at the expense of a substantial increase in pixel-level distortion. It remains unanswered whether one can attain similar improvements in the perceptual quality of images from the text, with only minimal degradation in PSNR.

\section{Method}\label{sec:method}

We introduce TACO (Text-Adaptive COmpression), a simple yet effective text-guided image compression algorithm that can achieve both high pixel-level and perceptual fidelity.

At a high level, TACO is a simple framework that transforms a conventional PSNR-oriented image compression backbone into a text-guided one. TACO first augments the encoder of the backbone with a text adapter module, which inserts the textual information into the encoder. Then we train the overall network from scratch using a joint image-text loss.

\subsection{Background: Autoencoder-based Codec} \label{ssec:background}

As the backbone architecture, we use ELIC \citep{elic}, a PSNR-oriented neural image compression codec. We use ELIC as a base model due to its popularity and performance, and TACO does not rely on any of its structural properties. In principle, we expect TACO to be able to be combined with other base models to enhance their perceptual quality.

Similar to most autoencoder-based codecs, ELIC roughly consists of four components: encoder, quantizer, entropy model, and decoder. An image is processed sequentially by the components as follows. First, the encoder maps an \textit{image} $x$ to a \textit{latent feature} $y$. The feature is then quantized into $\hat{y}$. The \textit{quantized feature} $\hat{y}$ is (losslessly) compressed into the code and decompressed with the entropy model. Finally, the decoder synthesizes the \textit{reconstruction} $\hat{x}$ from the decompressed feature $\hat{y}$.
\begin{align}
x \xrightarrow{\mathrm{encode}} y \xrightarrow{\mathrm{quantize\:\&\:entropy\:coding}} \hat{y} \xrightarrow{\mathrm{decode}} \hat{x}
\end{align}
The encoder, where TACO injects the text information, is a stack of three residual bottleneck blocks (RBs; \citet{he16}). The residual blocks are interleaved with five $5\times5$ convolutional layers (with stride $2$) and two attention modules \citep{cheng2020}. For a more detailed description, see the original paper of \citet{elic}.

\begin{figure}[t]
    \centering
    \includegraphics[width=\columnwidth]{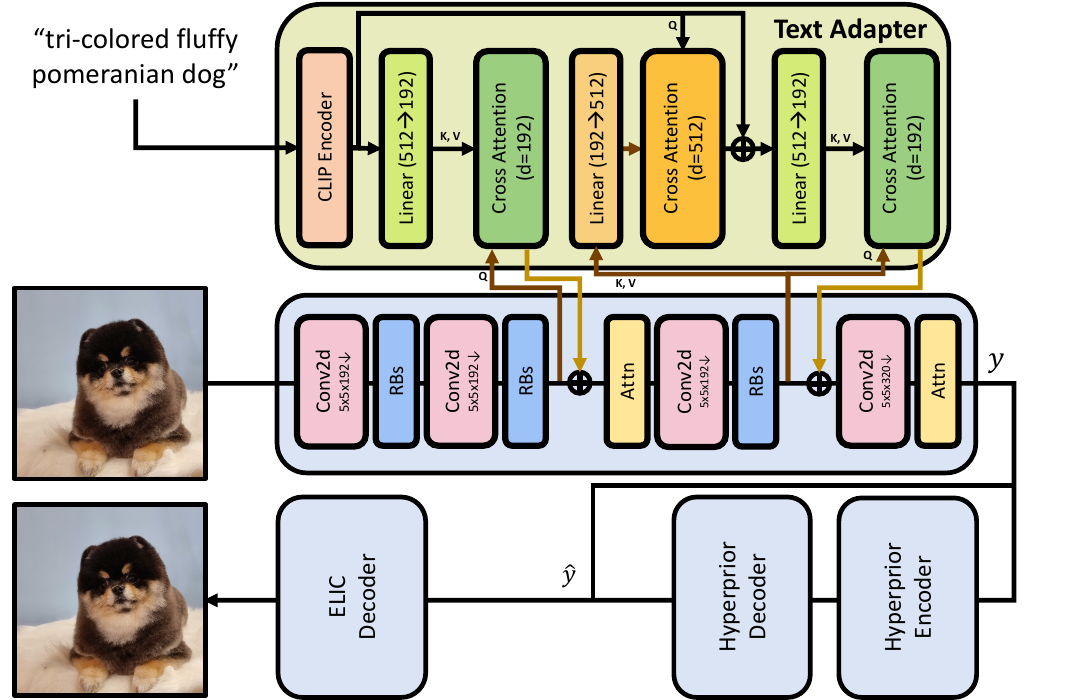}
    \caption{\textbf{Text adapter of TACO}. The text adapter first extracts features from the image caption using the CLIP text encoder. The textual features are then injected into the ELIC encoder through multiple cross-attention layers, interleaved with linear layers.}
    \label{fig:overall_archi}
\end{figure}

\subsection{TACO: Text-Adaptive Compression}\label{ssec:taco}

To enable an effective utilization of the textual information, TACO transforms the backbone image compression model in two ways. First, TACO injects the text information to the encoder of the backbone through \textit{text adapter}. Second, TACO trains the whole model (\ie, backbone model and the text adapter) with the combined loss function of rate, distortion, and the joint image-text loss.

\textbf{Text Adapter.} The text adapter takes in the text segment $c$ associated with the original image $x$ (\eg, image caption) as an input, and injects the text information into the backbone encoder to generate the joint image-text latent feature $y$.
\begin{align}
x \xrightarrow{\mathrm{encode}} y \quad \Rightarrow \quad (x,{\color{purple}c}) \xrightarrow{\mathrm{encode}\:{\color{purple}\mathrm{+\:TACO}}} y
\end{align}

The text adapter operates in two steps (\cref{fig:overall_archi}). (1) Embedding: Maps the given text to the joint image-text feature space. (2) Injection: Gradually processes and injects text features into the backbone encoder through multiple layers.

\textbf{Adapter: Embedding.} We leverage a pre-trained CLIP encoder \citep{clip}, which contains rich semantic information on image-text correspondences. By using CLIP features, we expect reductions in the amount of extra computation and data for further reducing the domain gap between features. We use CLIP as frozen without fine-tuning.

More concretely, this step generates a length-$m$ sequence of $d$-dimensional text tokens. In our experiments, we use $m=38$, the maximum token length of MS-COCO captions, and $d=512$, the dimensionality of CLIP encoder outputs.
\begin{align}
c \xrightarrow{\text{CLIP encoder}} \mathbf{e} := (e_1,\ldots,e_m), \quad e_i \in \mathbb{R}^d.
\end{align}

\begin{figure*}[!tb]
\centering 
    \includegraphics[width=0.85\textwidth ]{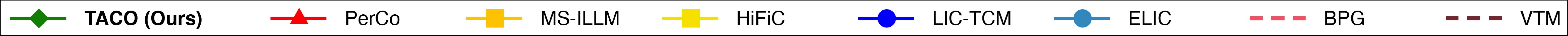}
    \vskip 0.1in
    \setlength{\tabcolsep}{1pt}
    \begin{tabular}{ccc}
    \includegraphics[width=0.33\textwidth ]{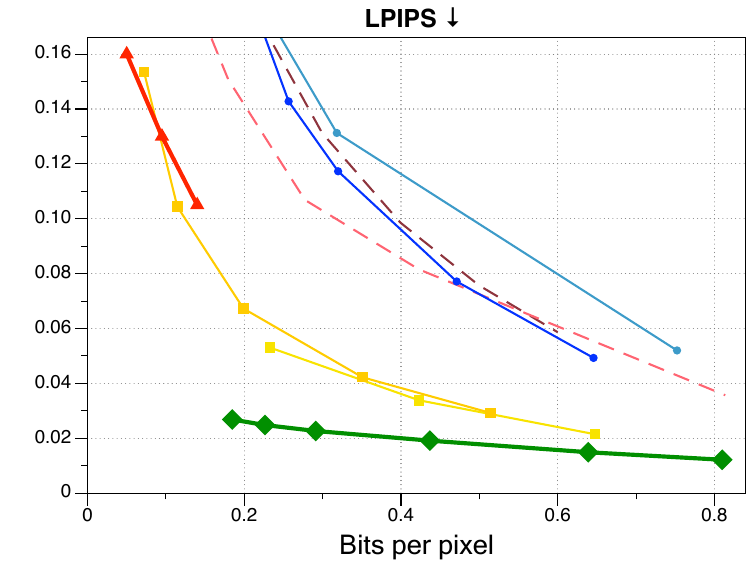}& \includegraphics[width=0.33\textwidth ]{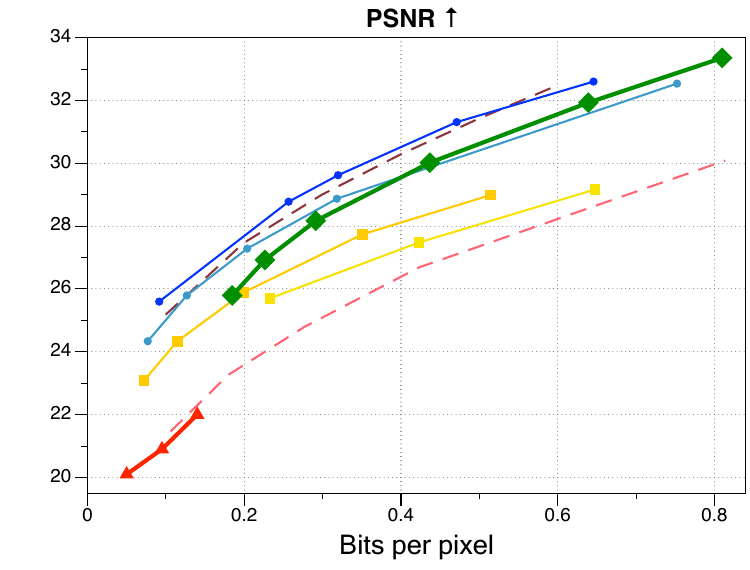}& \includegraphics[width=0.33\textwidth ]{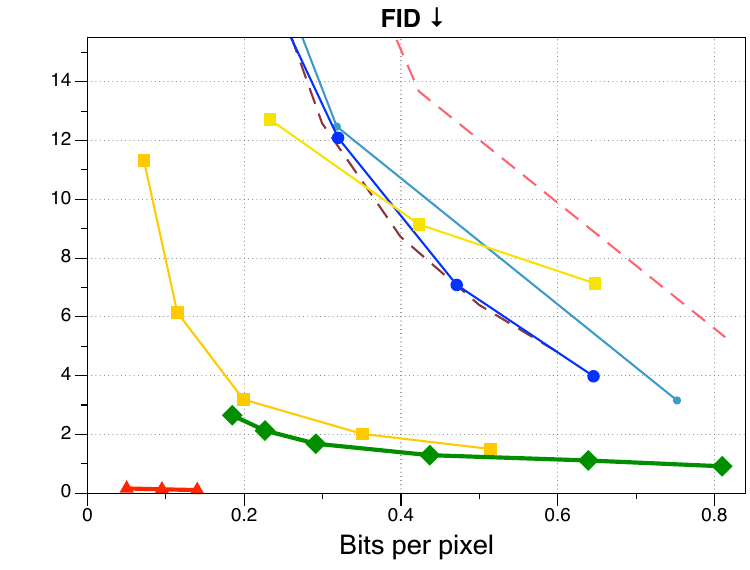} 
    \end{tabular}
    
    \caption{\textbf{Compression results: MS-COCO 30k.} TACO achieves the best or competitive results in all metrics. In particular, TACO achieves the best LPIPS among all methods considered, while achieving only $\sim$1dB less PSNR than LIC-TCM. PerCo achieves an impressive FID, but TACO outperforms in terms of both PSNR and LPIPS. 
    We also note a strange under-performance of HiFiC in FID under this setup. This may be due to the resizing operation when measuring FID.}
    \label{fig:result_coco30k}
\end{figure*}

\textbf{Adapter: Injection.} We use a six-layer network consisting of three linear layers and three cross-attention (CA) layers. Here, CAs play a crucial role injecting the text knowledge to images, or vice versa. This use of CA for knowledge injection is inspired by \citet{vitadapter}, who uses CA for adapting ViTs to another vision task (dense prediction). In this work, we repurpose CAs for injection across different modalities. Each CA plays the following role:
\begin{itemize}[leftmargin=*,topsep=-1pt,parsep=0pt,itemsep=0.5pt]
\item \textit{\textbf{CA1: Text$\to$Image.}} Updates image features based on text tokens. Given a $(h,w,k)$-dimensional image feature, we treat each spatial location as an image token. That is, the set of image tokens can be written as:
\begin{align}
\mathbf{v} = (v_1,\ldots,v_{h\times w}), \quad v_i \in \mathbb{R}^k.
\end{align}
CA1 computes query from each $v \in \mathbf{v}$ and keys and values from $e \in \mathbf{e}$. To match the dimensionality of $e$ and $v$, we put a linear layer before CA1. The output of the layer is
\begin{align}
\tilde{v}_i = \mathrm{LN}(v_i) + \gamma \odot \mathrm{Attention}(\mathrm{LN}(v_i),\mathrm{LN}(\mathrm{Lin}_1(\mathbf{e}))),
\end{align}
where $\mathrm{Attention}(\cdot)$, $\mathrm{LN}(\cdot)$, and $\mathrm{Lin}(\cdot)$ denotes the attention, LayerNorm, and linear layer, respectively. Here, $\gamma$ is a learnable weight hyperparameter.
\item \textit{\textbf{CA2: Image$\to$Text.}} Updates text tokens based on image features which are further processed by the backbone encoder. The operation is similar to CA1, but we compute queries from the text and keys/values from the image.
\item \textit{\textbf{CA3: Text$\to$Image (downsampled).}} Same operation as CA1, but with two changes: First,  the text tokens have been updated by image features (by CA2). Second, image features have been downsampled, so that text affects more global image features than in CA1.
\end{itemize}

\textbf{Joint Image-Text Loss.} To train the model, we use a mixture of four different loss functions. Given the original image $x$, the text description $c$, the reconstructed image $\hat{x}$, and its quantized latent feature $\hat{y}$, we use the loss
\begin{align}
r(\hat{y}) &+ \lambda \cdot d(x,\hat{x}) + k_{\mathrm{p}}\cdot \mathrm{LPIPS}(x,\hat{x}) + k_{\mathrm{j}} \cdot L_{\mathrm{j}}(x,\hat{x},c)
\end{align}
The first three losses are standard: $r(\cdot)$ denotes the rate, $d(\cdot,\cdot)$ denotes the MSE, and $\mathrm{LPIPS}(\cdot,\cdot)$ denotes the LPIPS loss. The fourth loss $L_{\mathrm{j}}(\cdot,\cdot,\cdot)$ is what we call the \textit{joint image-text loss}, which encourages the reconstructed image to be semantically close to the given text and the original image. Formally, the loss is defined as the following:
\begin{align}
    L_{\mathrm{j}}(x,\hat{x},c) = L_{\mathrm{con}}(f_{\mathrm{I}}(\hat{x}), f_{\mathrm{T}}(c)) + \beta \cdot \|f_{\mathrm{I}}(x) - f_{\mathrm{I}}(\hat{x})\|_2. \label{eq:jitloss}
\end{align}

Here, the functions $f_{\mathrm{I}}(\cdot),f_{\mathrm{T}}(\cdot)$ denote the CLIP image and text embeddings, and $L_{\mathrm{con}}(\cdot,\cdot)$ denotes the contrastive loss used in CLIP \citep{clip}. In other words, the joint image-text loss is a mixture of (1) the relevance between the text and reconstructed image and (2) the CLIP embedding distance between the original and reconstructed image. We note that \citet{tgic} also uses a loss similar to eq~\eqref{eq:jitloss}, but based on the AttnGAN embedding \citep{xu18}.

For better PNSR, we do not use the adversarial loss. Such practice is common in perception-oriented codecs, such as MS-ILLM or HiFiC, but not in PSNR-oriented codecs. TACO works by improving the perceptual quality of PSNR-oriented methods with texts instead of further improving the perceptual quality of already perceptually good models.

\section{Experiments}\label{sec:experiment}

\begin{figure*}[t]
    \centering 
    \includegraphics[width=0.7\textwidth ]{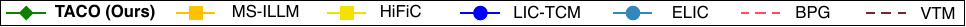}
    \vskip 0.1in
    \setlength{\tabcolsep}{1pt}
    \begin{tabular}{ccc}
    \includegraphics[width=0.32\textwidth ]{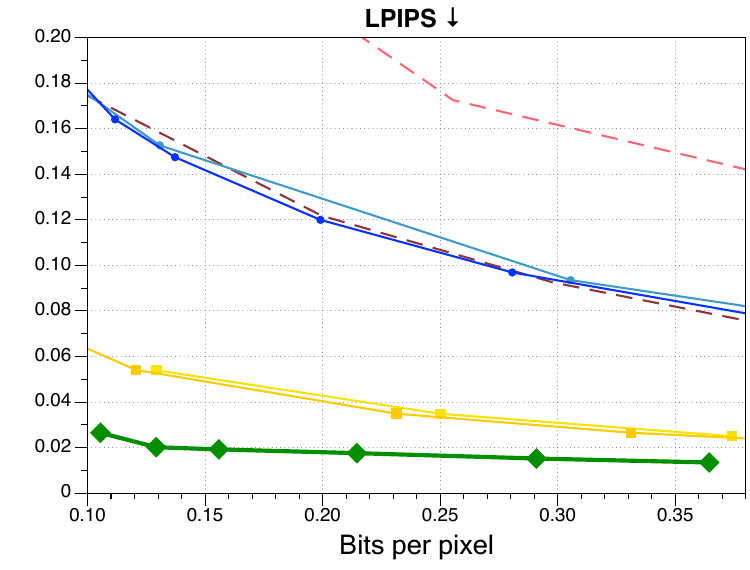}& \includegraphics[width=0.32\textwidth ]{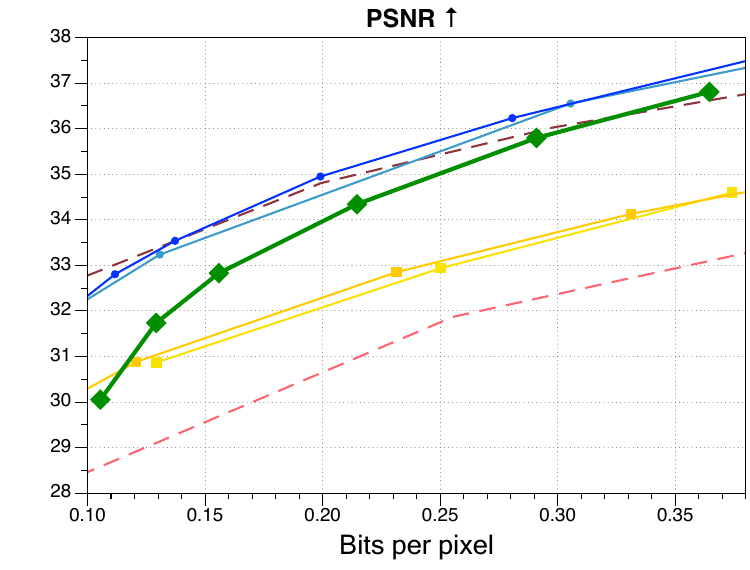}& \includegraphics[width=0.32\textwidth ]{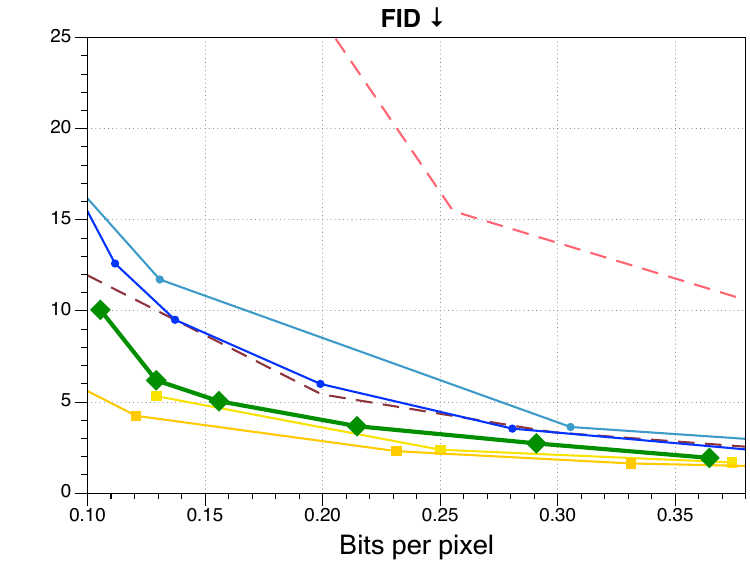}\\ 
    \end{tabular}
    \caption{\textbf{Compression results: CLIC.} TACO outperforms all other baselines in LPIPS, and achieves close-to-best results in PSNR and FID. In particular, TACO achieves the PSNR very close to LIC-TCM and ELIC, vastly outperforming realism-oriented codecs.}
    \label{fig:result_clic}
\end{figure*}

\begin{figure*}[!tb]
    \centering 
    \setlength{\tabcolsep}{1pt}
    \begin{tabular}{ccc}
    \includegraphics[width=0.32\textwidth ]{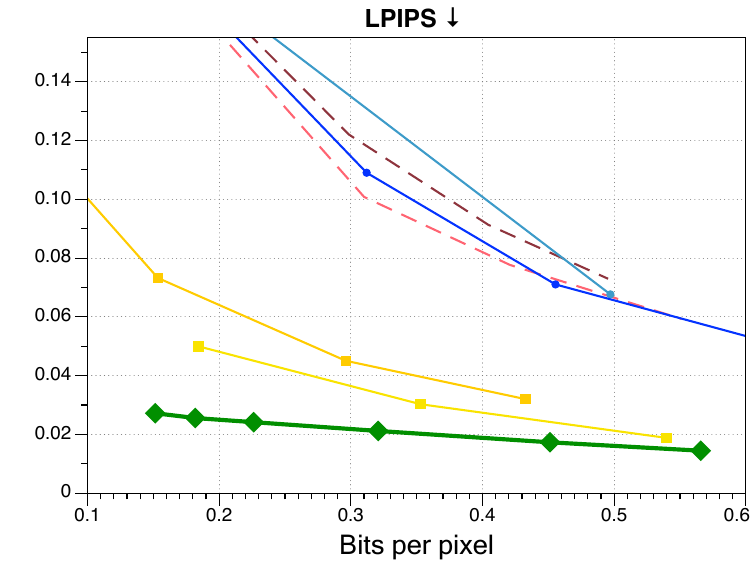}& \includegraphics[width=0.32\textwidth ]{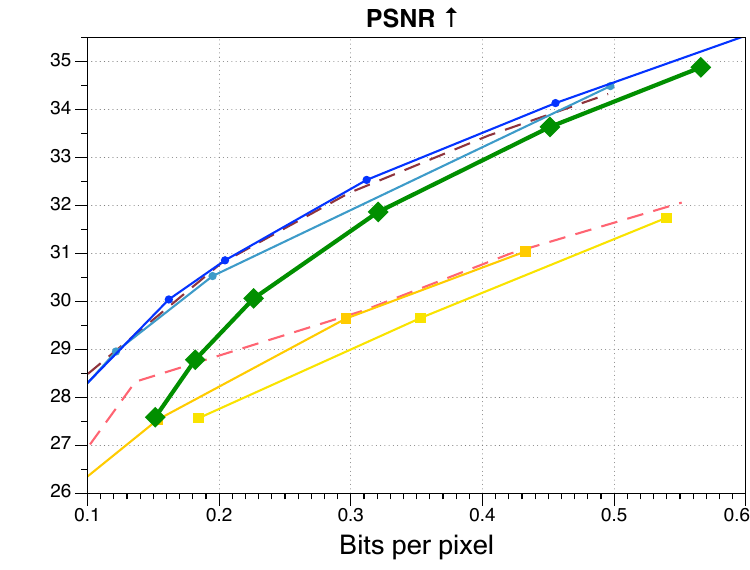}& \includegraphics[width=0.32\textwidth ]{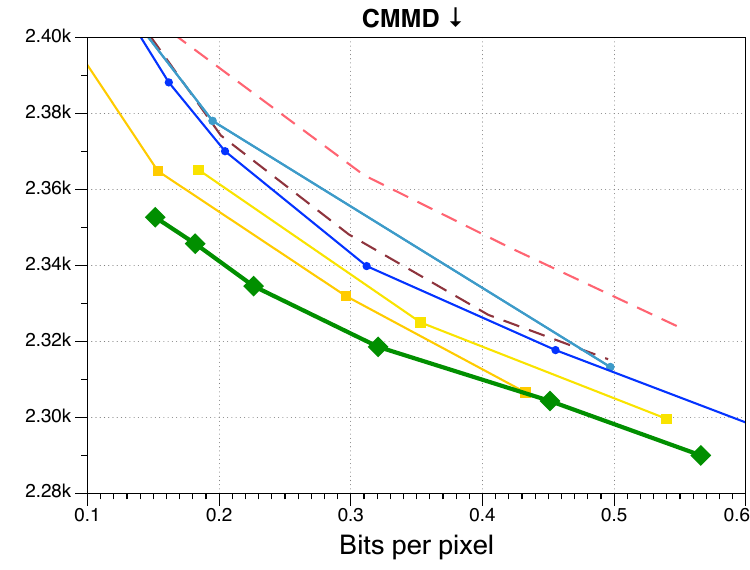}
    \end{tabular}
    \caption{ \textbf{Compression results: Kodak.} Similar to MS-COCO and CLIC, TACO outperforms baselines with a substantial margin in terms of LPIPs, and closely matches the PSNR of LIC-TCM and ELIC. TACO also performs best in terms of CMMD.}
    \label{fig:result_kodak}
\end{figure*}

This section is organized as follows. \cref{ssec:expsetup} describes the basic experimental setup. \cref{ssec:mainresults} compares TACO with image compression codecs that do not utilize text. \cref{ssec:comparison_textguided} compares with text-guided codecs; these works do not release code and/or model checkpoints and evaluation setups vary significantly from paper to paper, necessitating a more fine-grained comparison. Finally, \cref{ssec:ablations} provides various analyses and ablations on TACO.

\subsection{Experimental Setup}\label{ssec:expsetup}

\textbf{Datasets.} For training, we use the training split of the MS-COCO dataset \citep{mscoco}, which consists of 82,783 images with five human-annotated captions for each image; we use all five captions for training. We randomly crop each image to $256\times256$ resolution for training.

For evaluation, we use three different datasets. (1) MS-COCO 30k: We use the ``restval'' subset of the MS-COCO validation split, which consists of total 30,504 images. For evaluation, we center-crop the images to the $256\times256$ resolution; we use center drop to more faithfully preserve the semantic information in the image caption.\footnote{This is slightly different from what is called ``Dall-E processing'' \citep{dalle}, which have been used by \citet{multi_realism}, in that they perform random cropping.} (2) CLIC \citep{clic2020}: The test set of the Challenge on Learned Image Compression 2020, consisting of 428 images. We do not crop these images. (3) Kodak dataset \citep{kodak}: A natural image dataset that consists of $24$ images. We also evaluate on these images as-is. We note that CLIC and Kodak do not have any associated captions; we use the image captions generated by OFA \citep{ofa}.

\textbf{Metrics.}
We focus on three image quality metrics:
\begin{itemize}[leftmargin=*,topsep=-1pt,parsep=0pt,itemsep=0.5pt]
\item \textit{\underline{LPIPS}} is a perceptual fidelity metric that measures the distance between the original and reconstructed image in the deep feature space; we use AlexNet features.
\item \textit{\underline{PSNR}} is a pixel-wise fidelity metric, measured in dB.
\item \textit{\underline{FID}} is a realism metric that measures the statistical fidelity between two image distributions (original and reconstructed) of inception features. For measuring FID, we follow \citet{ms_illm} to resize the images to $299 \times 299$ resolution before measuring.
\end{itemize}
As a quantitative realism metric for Kodak dataset, we use a recently proposed CMMD \citep{cmmd} instead of FID. This is because FID is known to be unstable in small datasets. For CMMD, we use the publicly available version of CLIP (w.o. projection). We additionally report results on MS-SSIM, PieAPP, and CMMD in the  \cref{app:more_rd_results}.

\textbf{Optimization \& Hyperparameters.} We use Adam with batch size $4$, and train for $50$ epochs. The initial learning rate is set to $10^{-4}$, and we decay the learning rate by $1/10$ at epochs 45 and 48. For hyperparameters, we simply use a fixed set $(k_{\mathrm{p}},k_{\mathrm{j}},\beta) = (3.5,0.0025,40)$ throughout all bpps, instead of conducting an extensive hyperparameter tuning for each bpp. To get models of various bitrates, we train models with $\lambda \in \{4,8,16,40,90,150\} \times 10^{-4}$.

\begin{figure*}[t]
    \centering 
    \setlength{\tabcolsep}{1pt}
    \begin{tabular}{ccc}
    \includegraphics[width=0.33\textwidth]{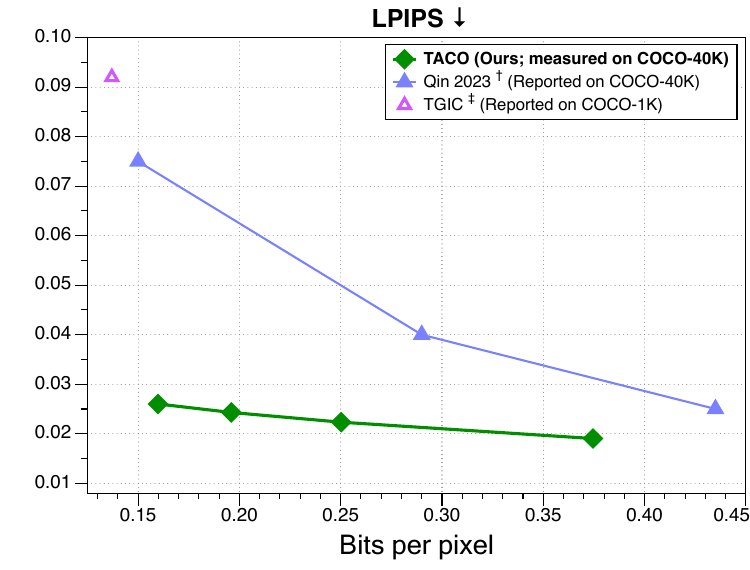}&      \includegraphics[width=0.33\textwidth]{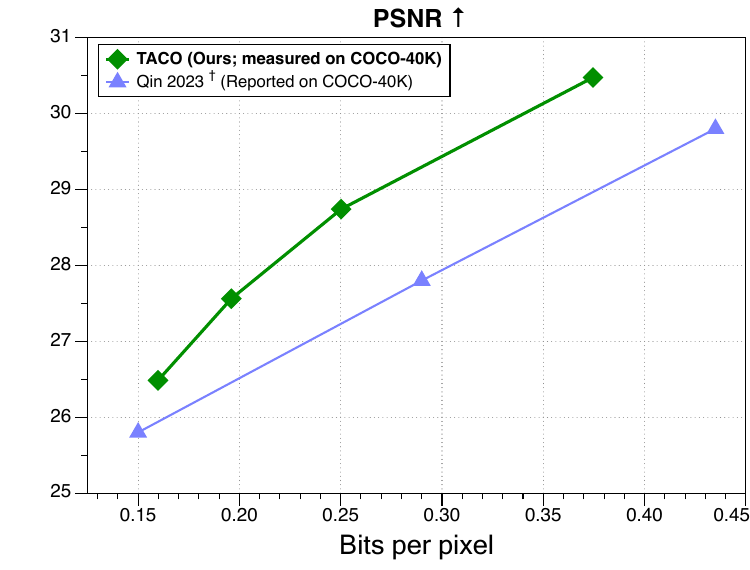}&       \includegraphics[width=0.33\textwidth]{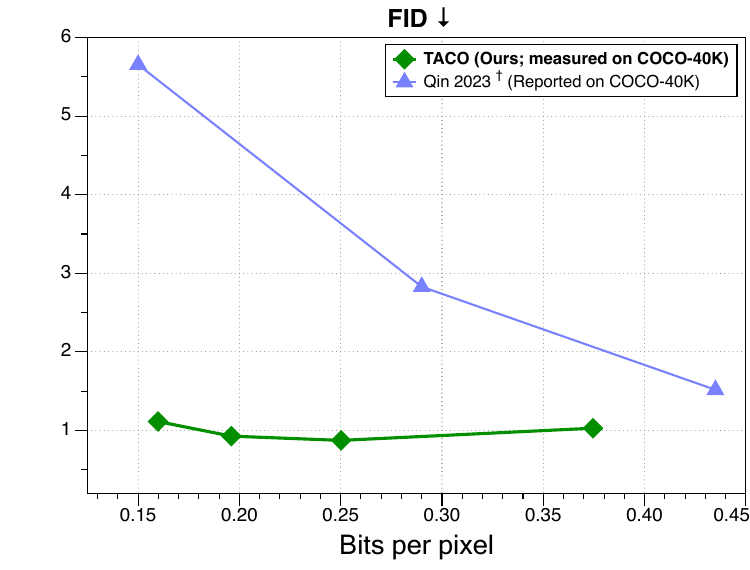}\\ 
    \end{tabular}
    \caption{\textbf{Comparison with text-guided decoding baselines.} We compare with the MS-COCO 40k results reported by \citet{qin_comp_23}, to the performance of TACO on the same dataset. We also compare with the MS-COCO 1k result reported by TGIC \citep{tgic}.}
    \label{fig:compare_with_text_guided_methods} 
\end{figure*}

% \begin{figure}[t]
%     \centerline{\includegraphics[width=0.32\columnwidth]{assets/figures/LPIPS_Comparison_with_PIC_TGIC.pdf}} 
%     % \centerline{\includegraphics[width=0.32\columnwidth]{assets/figures/PSNR_Comparison_with_PIC_TGIC.pdf}}
%     % \centerline{\includegraphics[width=0.32\columnwidth]{assets/figures/FID_Comparison_with_PIC_TGIC.pdf}}
%     \caption{\textbf{Comparison with text-guided baselines.} Qin 2023 reports compression performance on MS-COCO 40k, and TGIC reports performance on MS-COCO 1K.}
%     \label{fig:compare_with_text_guided_methods} 
% \end{figure}
% \begin{figure}[t]
%     % \centerline{\includegraphics[]{example-image-duck}} % 
%     \centerline{\includegraphics[width=\columnwidth]{assets/figures/Ablation_Study/uni_directional/Abl_Uni_LPIPS.pdf}} % 
%     \caption{\textbf{Ablation: Adapter structure.} We compare the efficacy of the proposed text adapter with the image-text attention module, used by \citet{tgic}. Our text adapter shows superior performance when attached on ELIC encoder.}
%     \label{fig:ablation_adapter} 
% \end{figure}

\begin{figure*}[t]
    \centering 
    \setlength{\tabcolsep}{1pt}
    \begin{tabular}{ccc}
    \includegraphics[width=0.33\textwidth]{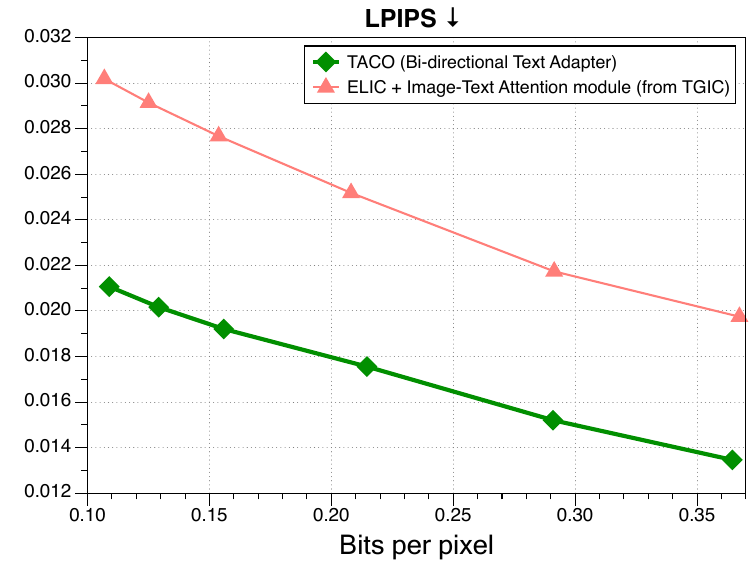}&      
    \includegraphics[width=0.33\textwidth]{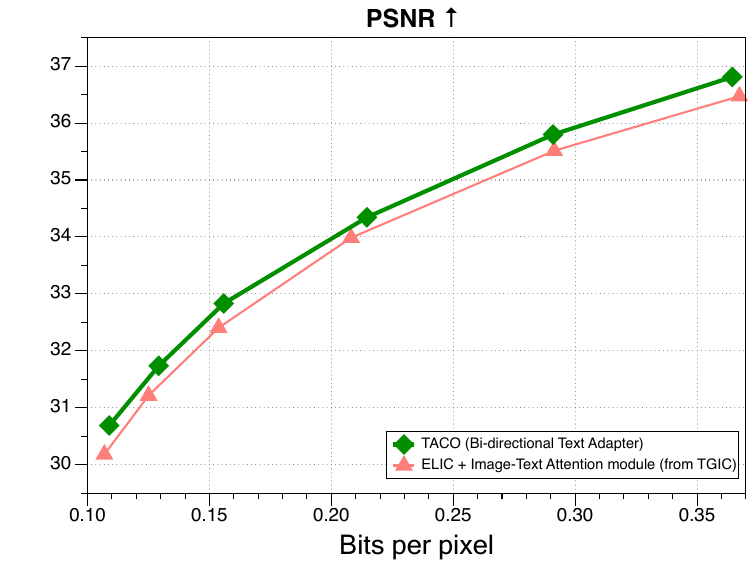}& \includegraphics[width=0.33\textwidth]{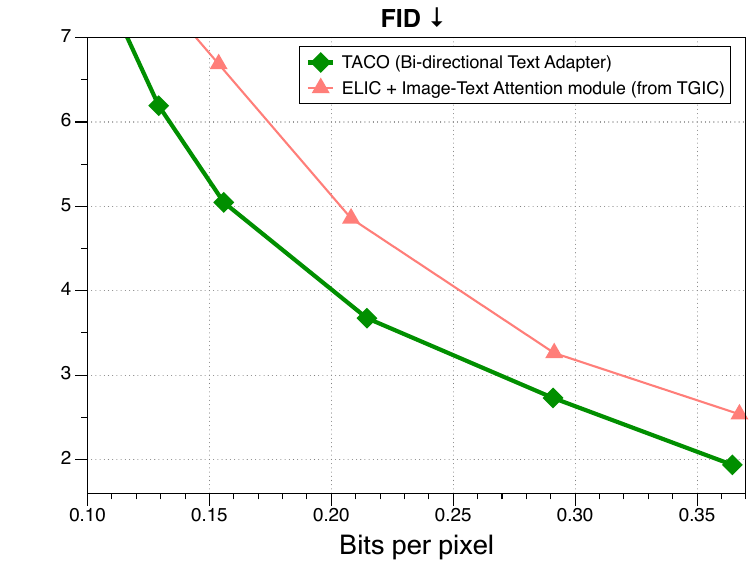}\\ 
    \end{tabular}
    \caption{\textbf{Text adapter vs. TGIC module.} We compare the effectiveness of the proposed text adapter with the image-text attention module, used by \citet{tgic}, on CLIC. Our text adapter performance better in all metrics, when attached on ELIC encoder.}
    \label{fig:ablation_adapter} 
\end{figure*}

\subsection{Results: vs. Image Compression Codecs}\label{ssec:mainresults}

We now compare the compression performance of TACO against various image compression codecs, including:
\begin{itemize}[leftmargin=*,topsep=-1pt,parsep=0pt,itemsep=0.5pt]
\item Handcrafted image compression codecs: VTM, BPG.
\item PSNR-oriented neural image compression codecs: ELIC \citep{elic}, and LIC-TCM \citep{lictcm}.
\item Perception-oriented neural codecs: HiFiC \citep{hific} and MS-ILLM \citep{ms_illm}.
\end{itemize}

We have used the official checkpoints for evaluation, except for ELIC (for which no official checkpoint is available).\footnote{We use the checkpoints from the following Github repo:\\ \href{https://github.com/VincentChandelier/ELiC-ReImplemetation}{https://github.com/VincentChandelier/ELiC-ReImplemetation}} We provide quantitative comparisons on MS-COCO 30k, CLIC, and Kodak on \cref{fig:result_coco30k,fig:result_clic,fig:result_kodak}, respectively.

\begin{figure*}[h]
    \centering 
    \setlength{\tabcolsep}{1pt}
    \begin{tabular}{ccc}
    \includegraphics[width=0.33\textwidth]{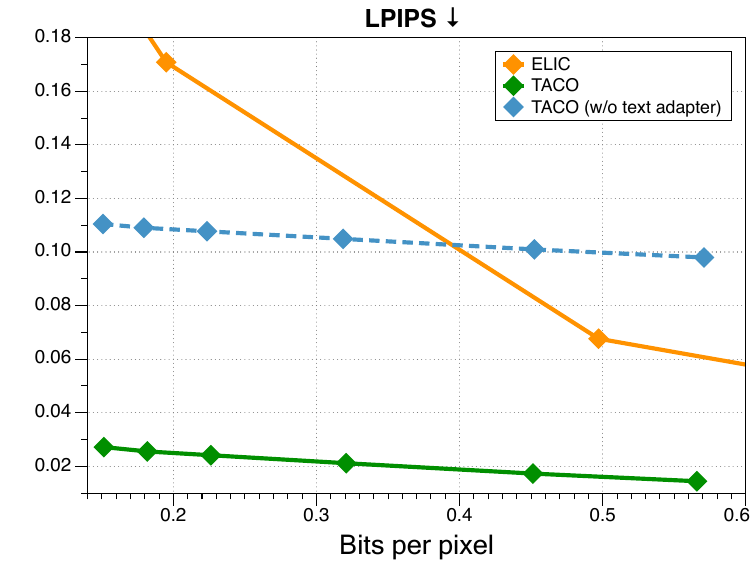}&      
    \includegraphics[width=0.33\textwidth]{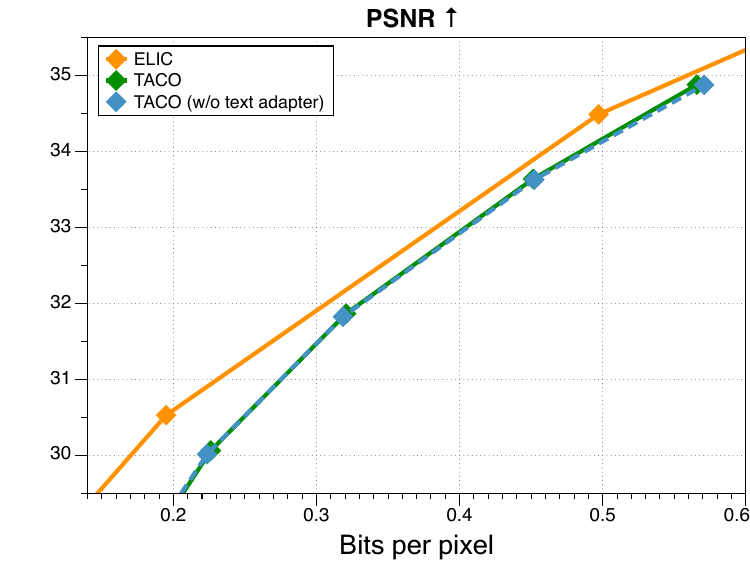}& \includegraphics[width=0.33\textwidth]{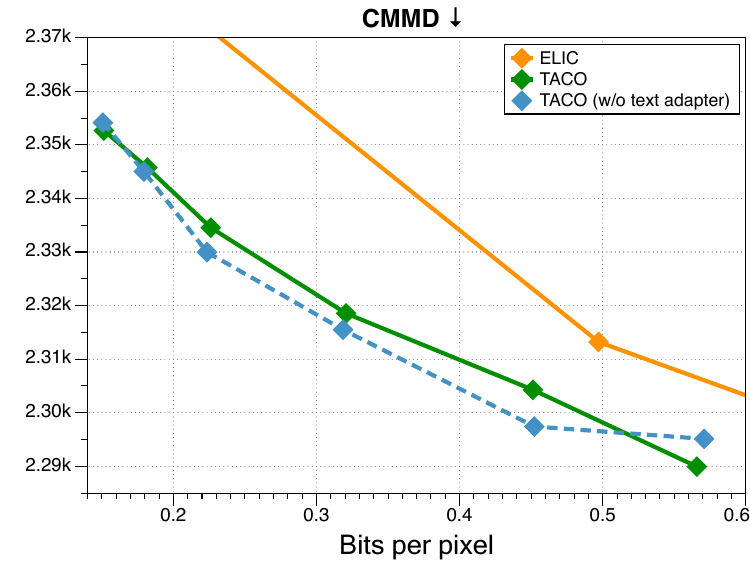}\\ 
    \end{tabular}
    \caption{\textbf{Without text adapter.} We observe both CMMD and LPIPS substantially degrade, while there is a tiny gain in PSNR.}
    \label{fig:without_text_adapter} 
\end{figure*}
\begin{figure*}[t]
    \centering 
    \setlength{\tabcolsep}{1pt}
    \begin{tabular}{ccc}
    \includegraphics[width=0.33\textwidth]{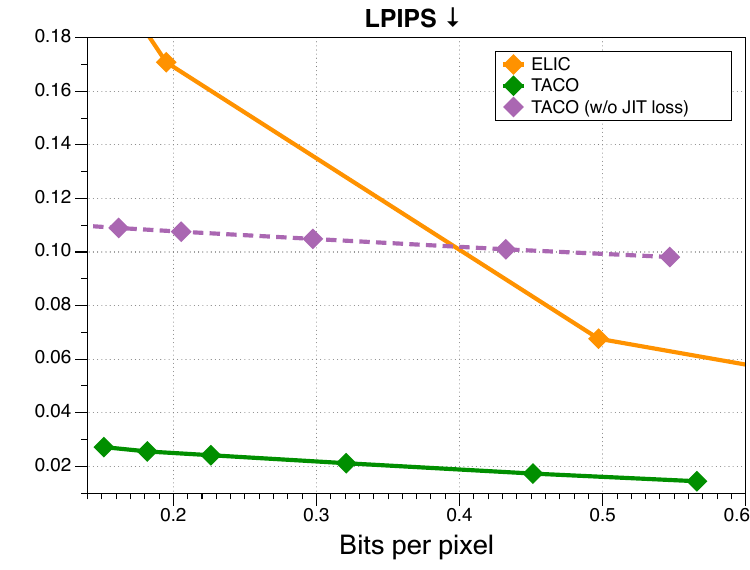}&      
    \includegraphics[width=0.33\textwidth]{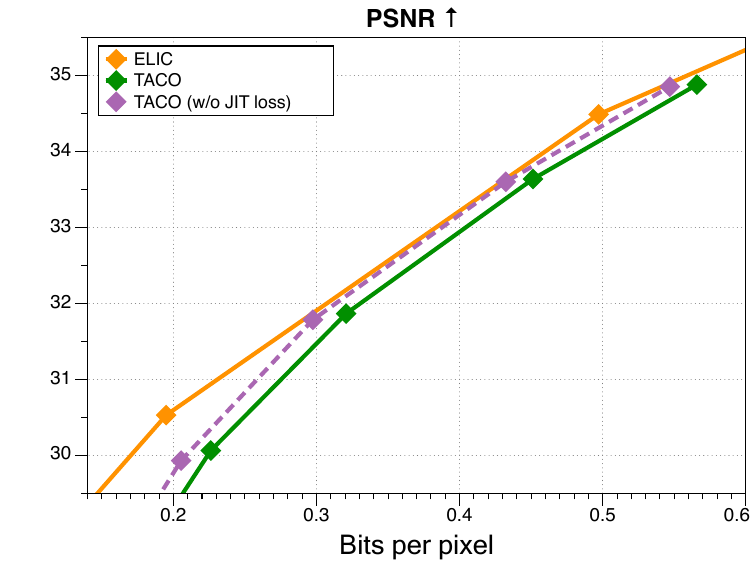}& \includegraphics[width=0.33\textwidth]{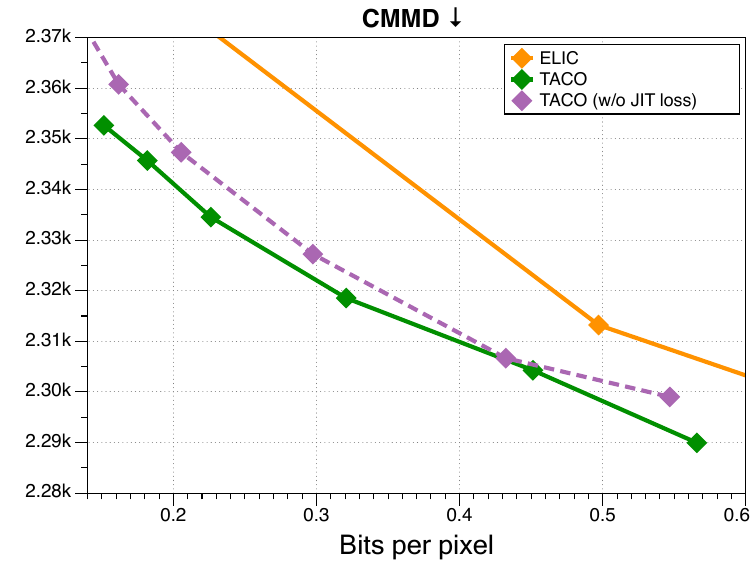}\\ 
    \end{tabular}
    \caption{\textbf{Without joint image text loss.} We observe LPIPS severely degrades, while PSNR remains similar, with a tiny gain in CMMD.}
    \label{fig:without_jit_loss} 
\end{figure*}

We observe that, in terms of LPIPS (perceptual fidelity), TACO outperforms all baselines in all datasets. For example, on MS-COCO 30k, TACO achieves 0.025 LPIPS at 0.226 bpp, which is less than half achieved by the best baseline (HiFiC), achieving 0.053 at 0.233 bpp. In terms of PSNR, TACO closely achieves the performance of ELIC and LIC-TCM, falling within the 1dB range from these baselines. In realism, TACO outperforms baselines in MS-COCO and Kodak, but not in CLIC; the gap In CLIC, however, tends to be very small, especially at a high bpp regime.

\subsection{Results: vs. Text-guided Decoding} \label{ssec:comparison_textguided}

We now compare the performance of TACO with text-guided codecs. In particular, we compare with codecs whose goal is to achieve high LPIPS at non-extreme bpp (\ie, over 0.1 bpp), such as TGIC \citep{tgic} and ``Qin 2023'' \citep{qin_comp_23}. Two things make a direct comparison difficult: (1) The baselines do not release official code or have model checkpoints publicly available. (2) The baselines report performance on non-unified setups: TGIC reports performance on MS-COCO 1k (along with CUB and Oxford-102), consisting of 1k randomly drawn test set samples. \citet{qin_comp_23} reports performance on MS-COCO 40k, \ie, the whole validation set, without any cropping. 

We focus on comparing with (more reproducible) Qin 2023 baseline. We evaluate TACO on MS-COCO 40k and compare it with the results reported in \citet{qin_comp_23}. From \cref{fig:compare_with_text_guided_methods}, we observe that TACO achieves much better performance than the previous text-guided baseline, based on the text-guided decoding strategy. This is somewhat surprising, as the original intention of using text-guided encoding has been on preventing the degradations in PSNR; instead, we also achieve better LPIPS and FID. This improvement can be attributed to the effectiveness of the text-adapter-based knowledge injection strategy.

\subsection{Ablation Studies} \label{ssec:ablations} 
To validate that both our proposed text adapter and joint image-text loss improve the performance, we conduct an ablation study. Through experiments on the Kodak dataset, we confirm that both components play essential roles.

\textbf{Without text adapter.} From \cref{fig:without_text_adapter}, we observe that perceptual fidelity greatly decreases without a text adapter (more than 4$\times$ in LPIPS). This gap is well-aligned with our intuition that the textual information plays an important role in how human perceives visual signals.

\textbf{Without joint image-text loss.} From \cref{fig:without_jit_loss}, we observe that the joint image-text loss contributes in improving not only the perceptual fidelity, but also realism (with a tiny degradation in PSNR). We hypothesize that the manifold-level information transferred from the CLIP embedding, which is trained with a large-scale dataset, helps the reconstructed image to remain realistic.

\subsection{Further Analyses} \label{ssec:analyses} 

We now ask ourselves several questions regarding the validity and practicality of the proposed method, TACO.

\textbf{Q1. Is the adapter architecture (\cref{ssec:taco}) effective?}\\
\textbf{A. Yes.} It outperforms the baseline method for text injection. We compare it to another text injection mechanism introduced in TGIC \citep{tgic}, which utilizes several convolutional layers to inject text information into the image encoder (coined ``image-text attention module''). We compare the performance of these two modules for injecting text information to the encoder, in \cref{fig:ablation_adapter}. From the figure, we observe that the proposed text adapter is indeed much more effective than what TGIC uses. 

\textbf{Q2. Does TACO preserve textual information better?}\\
\textbf{A. Yes, and No.} TACO-compressed images tend to preserve textual information much better than PSNR-focused methods, similar to LIC-TCM. We measure various text similarity scores between machine-generated captions for the original and reconstructed image (\cref{tab:textscores}); we use OFA for this purpose. We find that the images compressed by TACO preserve text semantics better than LIC-TCM and ELIC. On the other hand, the comparison with MS-ILLM is inconclusive; TACO performs slightly worse, but with slightly less bpp. In general, we find that models with high perceptual quality tend to preserve textual information better than PSNR-oriented compression codecs.

\begin{table}[t]
\caption{\textbf{Comparing text similarity.} We compare the text similarity scores for the CLIC images compressed by TACO and other baseline codecs. We find that TACO achieves much higher text similarity score than LIC-TCM, but is slightly worse than MS-ILLM that uses slightly more bits per pixels.}
\label{tab:textscores}
\centering
% \vspace{0.5em}
\resizebox{.48\textwidth}{!}{%
\begin{tabular}{lrrr}
    \toprule
     & BLEU-4 (↑) & CIDEr-D (↑) & SPICE (↑) \\
     & \scriptsize\citep{bleu} & \scriptsize\citep{cider}  & \scriptsize\citep{spice} \\
    \midrule
    LIC-TCM {\scriptsize (0.2810)} & 0.76 & 7.44 & 0.80 \\
    MS-ILLM {\scriptsize (0.2314)} & 0.85 & 8.39 & 0.87 \\
    ELIC {\scriptsize (0.3055)} & 0.78 & 7.58 & 0.81 \\
    \midrule
    \textbf{TACO} {\scriptsize (0.2146)} & 0.84 & 8.21 & 0.86 \\
    \bottomrule
\end{tabular}
}
\end{table}
\begin{table}[t]
\caption{\textbf{Computational cost of TACO.} We compare the wall-clock encoding and decoding latency of TACO with baseline image compression codecs. We average over 100 MS-COCO images (256$\times$256), randomly drawn from the validation split (0.0148 bpp). We also compare the computational cost when encoding a high-resolution image (`High' in table), observing that the increasing cost of attention remains relatively small (+10.2\% $\rightarrow$ +4.8\%).}
\label{tab:latency_comparison}
\resizebox{.48\textwidth}{!}{%
% \begin{center}
% \begin{small}
%\begin{sc}
\begin{tabular}{lrrrr}
    \toprule
     & Enc. (ms) &  Enc.@High (ms) & Dec. (ms) & Total (ms) \\
    \midrule
    LIC-TCM & 112.07 & 960.99 & 125.26 & 237.33 \\
    MS-ILLM & 70.39 & 800.45 &53.74 & 124.14 \\
    ELIC & 71.35 & 793.41 &102.07 & 173.42 \\
    \midrule
    \textbf{TACO} &  78.60 (+10.2\%) & 832.07 (+4.8\%) & 102.98 & 181.58 \\
    \bottomrule
\end{tabular}
%\end{sc}
% \end{small}
% \end{center}
}
\end{table}

\begin{table}[h]

\caption{\textbf{Memory efficiency of TACO.}
We compare the model parameter sizes of the TACO and baselines. TACO shows superior memory efficiency than perception-oriented codecs.}

\label{app:params_number}
\begin{center}
\begin{small}
%\begin{sc}
\begin{tabular}{lr}
    \toprule
    Modules &  Parameters (M) \\
    \midrule
    ELIC & 36.93 \\
    LIC-TCM & 45.41 \\
    HiFiC/MS-ILLM & 181.72 \\
    \midrule
    \textbf{TACO} & 101.75 \\
    \bottomrule
\end{tabular}
%\end{sc}
\end{small}
\end{center}
\end{table}
\begin{table}[t]
\caption{\textbf{Caption dependency.} We compare TACO reconstruction results with four different captions generation methods, spanning from human to GPT-4. Human and GPT-4 achieves the best results when measured on 100 random samples of MS-COCO, but other captioning methods work very competitively (0.0148 bpp). Underlined denotes the best result.}
\label{tab:caption_type}
\centering
% \vspace{0.5em}
\resizebox{.48\textwidth}{!}{%
%\begin{center}
%\begin{small}
%\begin{sc}
\begin{tabular}{lrrrr}
    \toprule
     & Human & OFA & BLIP-2 & GPT-4 \\
     & \scriptsize\citep{coco_captions} & \scriptsize\citep{ofa} & \scriptsize\citep{blip2} & \scriptsize\citep{gpt4}\\
    \midrule
    LPIPS & \underline{0.0435} & \underline{0.0435} & \underline{0.0435} & \underline{0.0435}\\
    PSNR & \underline{27.42} & 27.41 & 27.41 & \underline{27.42} \\
    \bottomrule
\end{tabular}
%\end{sc}
%\end{small}
%\end{center}
}
\end{table}

\textbf{Q3. How much computation does TACO add?}\\ % ---around 10\%
\textbf{A. $\sim$ 5--10\%, depending on the image resolution.} We compare the wall-clock encoding and decoding speed of the proposed TACO with the baselines in \cref{tab:latency_comparison}. We have measured the latency on a Linux GPU server equipped with one NVIDIA GeForce RTX 3090 and AMD EPYC 7313 16-Core CPU. From the table, we find that TACO indeed introduces some latency in compression speed. In particular, we observe that the encoding time increases by 7ms over the vanilla ELIC. We note that the overhead, however, is not very big. TACO still runs much faster than LIC-TCM and remains comparable with MS-ILLM. Most of the overhead comes from the CLIP embedding step, which can be effectively parallelized with image encoding steps. Furthermore, we have used the smallest version of CLIP, which can be run fast on conventional GPU devices. We also compare the encoding speed when using a high-resolution image (1788$\times$2641, second column in the \cref{tab:latency_comparison}). As a result, the computational cost of attention remains relatively small and we find that the computational cost of CLIP remains constant with respect to the resolution.

\textbf{Q4. How many parameters are used in TACO?}\\
\textbf{A. Not too big---smaller than perception-oriented codecs.} We compare the number of models in \cref{app:params_number}. TACO adds 64.82M parameters to the ELIC, where 63.17M comes from the CLIP and 1.65M comes from the text adapter. In total, TACO has 101.75M parameters, which requires $\sim$400MBs to be loaded on memory. While this is larger than vanilla ELIC, it is $\sim$ 80\% smaller than other perception-oriented codecs, such as MS-ILLM, and HiFiC.

\textbf{Q5. Can we use other captioning models?}\\
\textbf{A. Yes.} In \cref{tab:caption_type}, we compare the compression quality of TACO utilizing captions generated by various methods. We observe that GPT-4 achieves a human-like result, but other captioning models perform very competitively. From this observation, we conjecture that the compression quality of TACO is less dependent on its writing style, but more on the core content of the sentences.

\section{Conclusion}\label{sec:conclusion}
In this paper, we have studied how one can utilize auxiliary text information when compressing images. We find that, in order to achieve high PSNR as well as high perceptual quality, utilizing the text to generate better codes is a simple yet effective strategy; using text to guide the decoding procedure may be much more challenging task.

\textbf{Limitations and future work.} There are several limitations of the proposed framework, which we hope to address in the future work. First, the additional encoding cost for handling the text scales quadratically with respect to the sequence length. This is due to the design of our text injection mechanism which computes cross-attention between image and textual features. In the cases where we expect the text sequence to be very long, \eg, video-like data with voice transcripts as text, the encoding time can get prohibitively long. Another limitation is the scalability of training. Currently, we train TACO using the image-text dataset, which is substantially scarcer than image-only or text-only dataset. To overcome this problem, using a captioning model to generate image-text pairs for training may be a good baseline.

\section*{Acknowledgements}\label{sec:acknowledgement}
This work was supported by Samsung Advanced Institute of Technology and Institute of Information \& communications Technology Planning \& Evaluation (IITP) grant funded by the Korea government(MSIT) (RS-2019-II191906, Artificial Intelligence Graduate School Program(POSTECH)). We would like to thank Sadeep Jayasumana for helpful comment on implementing the CMMD. 

\section*{Impact Statement}\label{sec:acknowledgement}
Our paper aims to advance the general field of machine learning and data compression. We thus expect many societal consequences to follow. However, we do not feel any specific consequence should be highlighted further.

%Authors are required to include a statement of the potential broader impact of their work, including its ethical aspects and future societal consequences. This statement should be in a separate section at the end of the paper (co-located with Acknowledgements, before References), and does not count toward the paper page limit. In many cases, where the ethical impacts and expected societal implications are those that are well established when advancing the field of Machine Learning, substantial discussion is not required, and a simple statement such as: 

%“This paper presents work whose goal is to advance the field of Machine Learning. There are many potential societal consequences of our work, none which we feel must be specifically highlighted here.”

%The above statement can be used verbatim in such cases, but we encourage authors to think about whether there is content which does warrant further discussion, as this statement will be apparent if the paper is later flagged for ethics review.

\bibliographystyle{icml2024}
\bibliography{refs}

\newpage
\appendix
\onecolumn

\section*{Appendix}
\renewcommand\thefigure{A\arabic{figure}} 
\setcounter{figure}{0} 

\section{Additional Experimental Details} 
\textbf{Implementation of TACO.} As illustrated in \cref{fig:overall_archi}, the base model we use is a version of ELIC, which has an intermediate feature dimension of 192 and a final feature map dimension of 320. The CLIP encoder we use is the ``CLIPTextModel'' with a version of 
 `openai/clip-vit-base-patch32,' available through the HuggingFace. We have set the token length to be 38, therefore getting the text embedding sequences with the size of \(\mathbb{R}^{38\text{x}512}\) when inputting the caption to the CLIPTextModel. 
 
\textbf{Baselines.} As mentioned in the main text, for neural codecs, we mostly use the official model checkpoints. For BPG, we have used version 0.9.8. For VTM, we have used version 19.2.

\section{Additional Metrics on Compression Results} \label{app:more_rd_results}

We additionally report more compression results, MS-SSIM, PieAPP, and CMMD scores, measured on the MS-COCO 30k, CLIC, and Kodak datasets.
\begin{figure*}[!h]
    \centering 
    \includegraphics[width=0.7\textwidth ]{assets/figures/Legends.pdf}
    \setlength{\tabcolsep}{1pt}
    \begin{tabular}{ccc}
    \includegraphics[width=0.32\textwidth ]{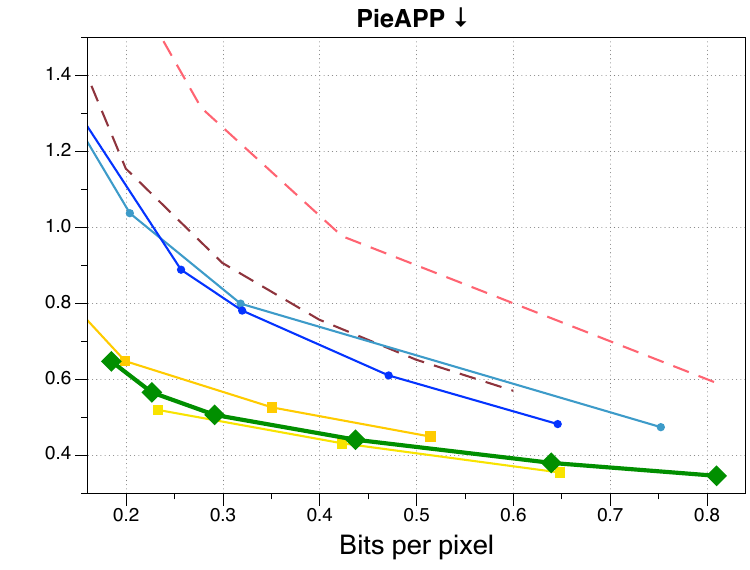}& \includegraphics[width=0.32\textwidth ]{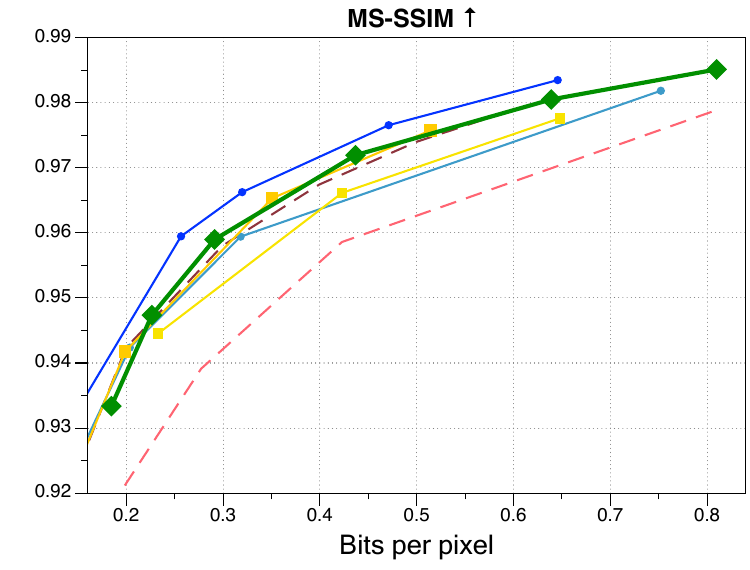}&
    \includegraphics[width=0.32\textwidth ]{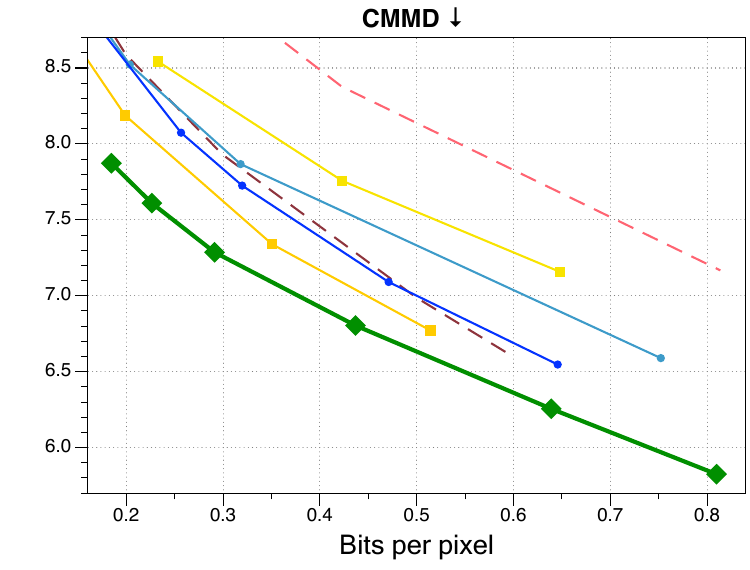}
    \end{tabular}
    % \caption{\textbf{MS-COCO 30k}}
    \label{fig:Appen_result_coco}
\end{figure*}

\begin{figure*}[!h]
    \centering 
    \setlength{\tabcolsep}{1pt}
    \begin{tabular}{ccc}
    \includegraphics[width=0.32\textwidth ]{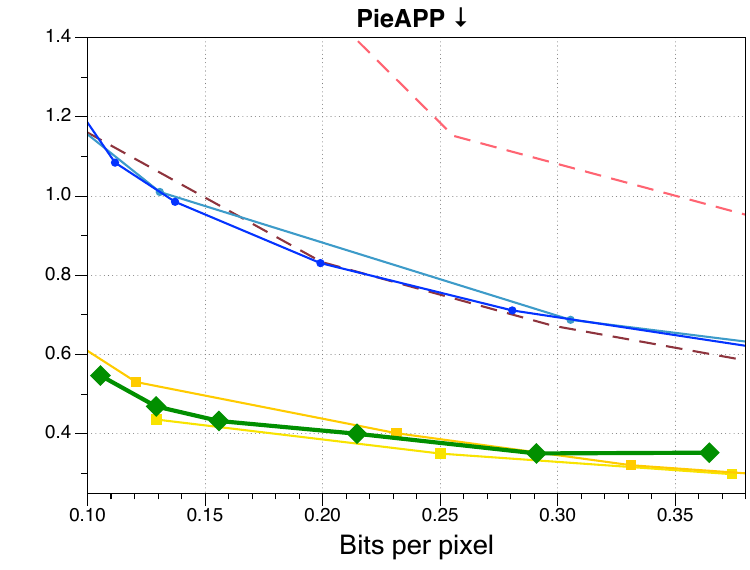}& \includegraphics[width=0.32\textwidth ]{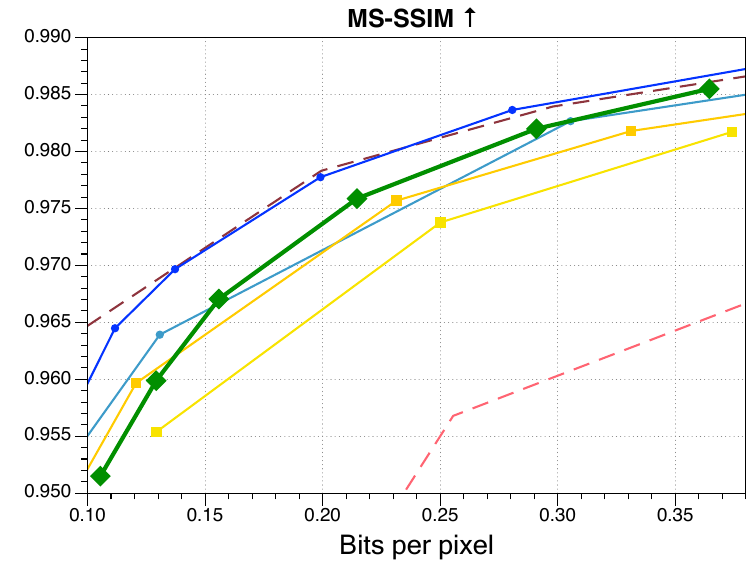}&
    \includegraphics[width=0.32\textwidth ]{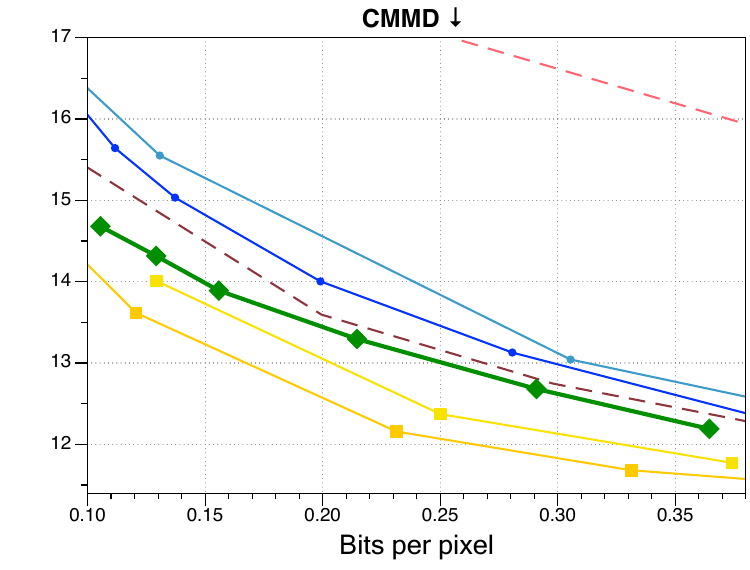}
    \end{tabular}
    % \caption{CLIC 2020}
    \label{fig:Appen_result_clic}
\end{figure*}

\begin{figure*}[!h]
    \centering 
    \setlength{\tabcolsep}{1pt}
    \begin{tabular}{ccc}
    \includegraphics[width=0.32\textwidth ]{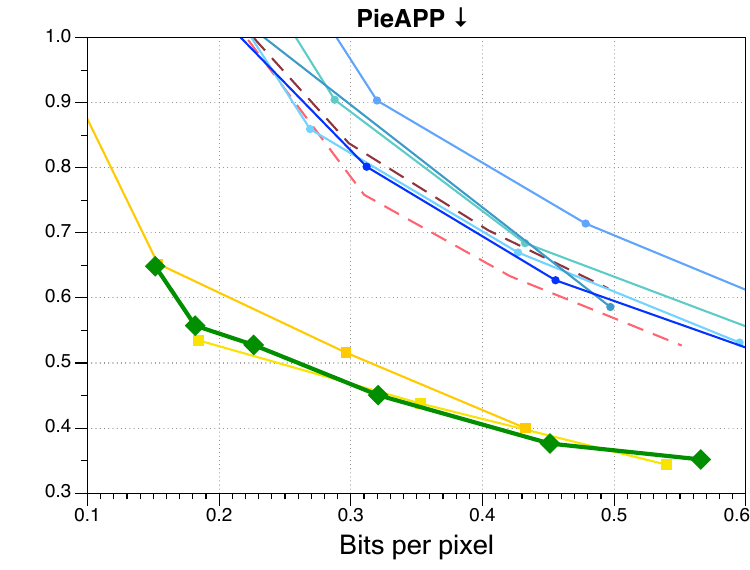}& \includegraphics[width=0.32\textwidth ]{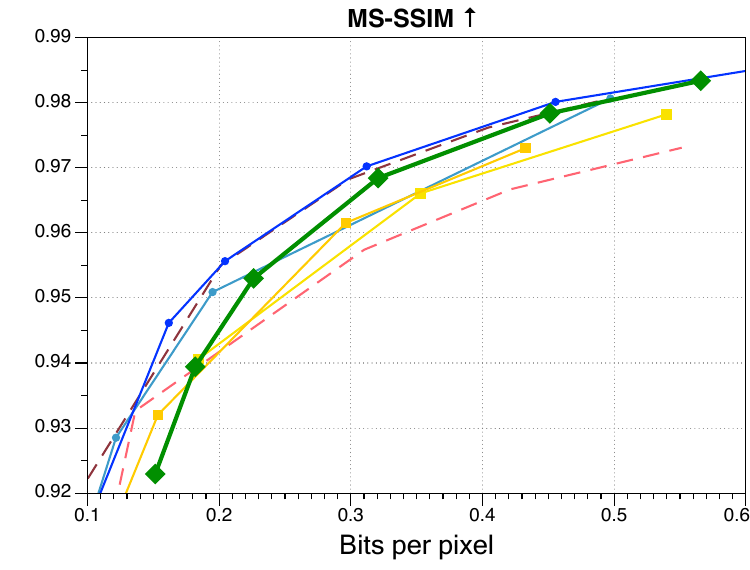}
    \end{tabular}
    \caption{\textbf{Additional quality metrics result on: MS-COCO 30k (\textit{top}), CLIC (\textit{middle}), Kodak (\textit{bottom})}}
    \label{app: Appen_result_kodak}
\end{figure*}

\section{Additional Ablation Study} 
\begin{figure}[h]
    \centering
    \setlength{\tabcolsep}{1pt}
    \begin{tabular}{ccc}
     \includegraphics[width=0.32\textwidth ]{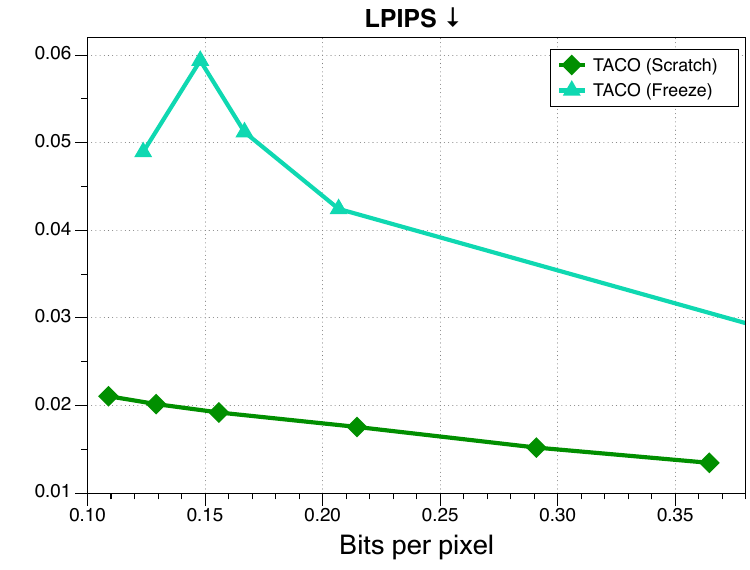}
    &
    \includegraphics[width=0.32\textwidth ]{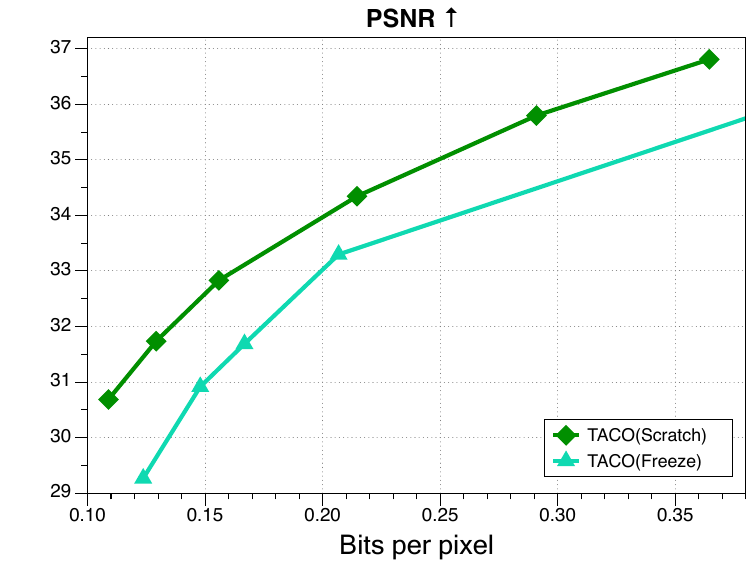}
    &
    \includegraphics[width=0.32\textwidth ]{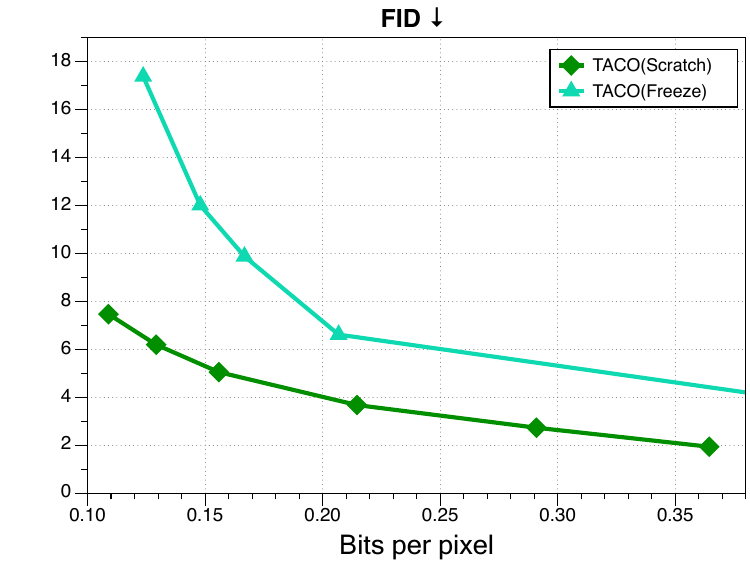}
    \\
    \end{tabular}
    \caption{Ablation result: freezing base model and only training text adapter}
    \label{fig:freeze_ablation}
\end{figure}

\textbf{Freezing base model.} We test whether freezing the base model (ELIC) to be a fully image-trained model is beneficial for performance or not. We find that this is not true (see \cref{fig:freeze_ablation}).

% {\textit{(top)}} 
\section{Additional Qualitative Results}\label{app:add_qualitative_examples}
We provide additional visualizations of the compression results of TACO and baselines in \cref{fig:qualitative_examples}. As has been observed in \cref{fig:qualitative}, we confirm that TACO suffers from fewer compression artifacts than MS-ILLM while reconstructing sharper details than LIC-TCM. We find that TACO consistently provides high-quality reconstructions, while LIC-TCM tends to overly smooth out the textures (see, e.g., the white wall under the cat) and MS-ILLM hallucinates small details (see, e.g., eyes of the cat).

\begin{itemize}[leftmargin=*,topsep=0pt,parsep=0pt,itemsep=0pt]
\item In the top display, we draw attention to two features: ``the reflections on glasses,'' and ``the texture of eyebrow and hair.'' MS-ILLM tends to hallucinate flat reflections on the glasses, and LIC-TCM tends to generate blurry eyebrow and hair.
\item In the middle display, we focus on the black metal staircase behind the net. MS-ILLM generates wavy textures on the metal, and LIC-TCM removes the net in front of the staircase. TACO also generates some hallucinative patterns but to a lesser degree. Also, we note that TACO is the only compression method that reconstructs (any) stud on the staircase.
\item In the bottom display, we highlight the gold hinge. MS-ILLM smooths out the details on the hinge, and LIC-TCM generates a blurry image overall. 
\end{itemize}

\begin{figure*}[!h]
    \centering 
    {\includegraphics[width=0.87\columnwidth]{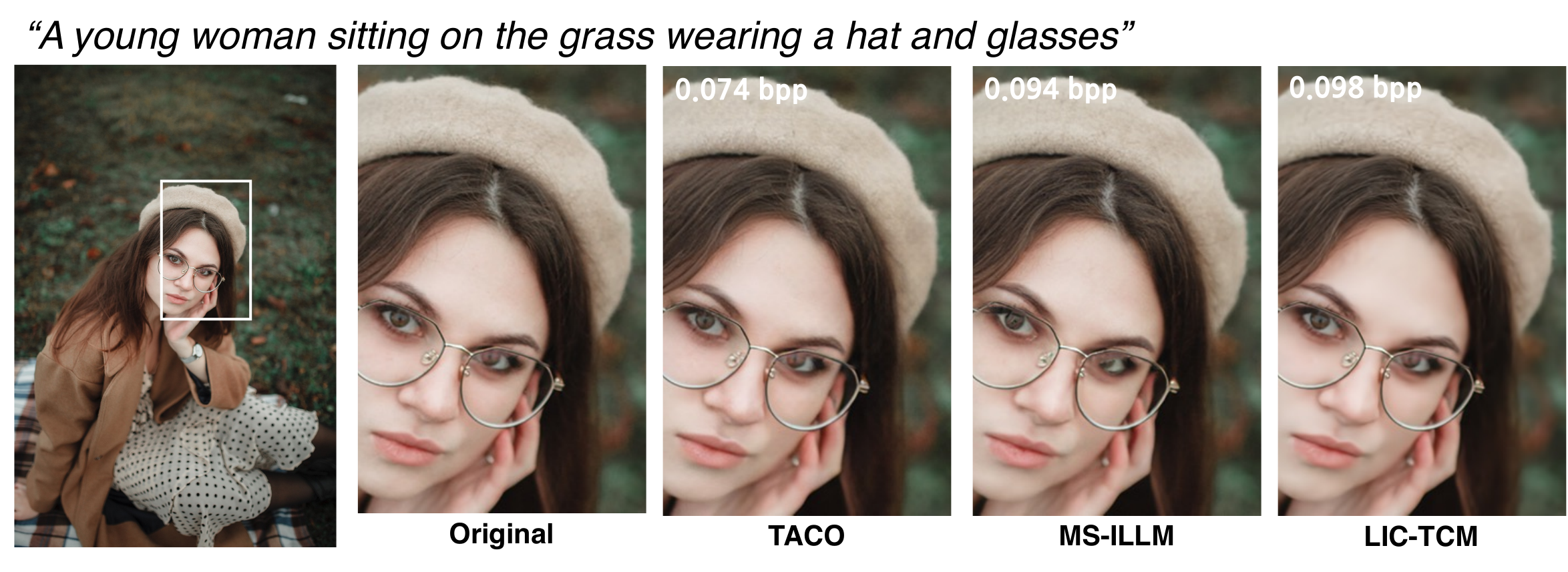}}\\      
    {\includegraphics[width=0.87\columnwidth]{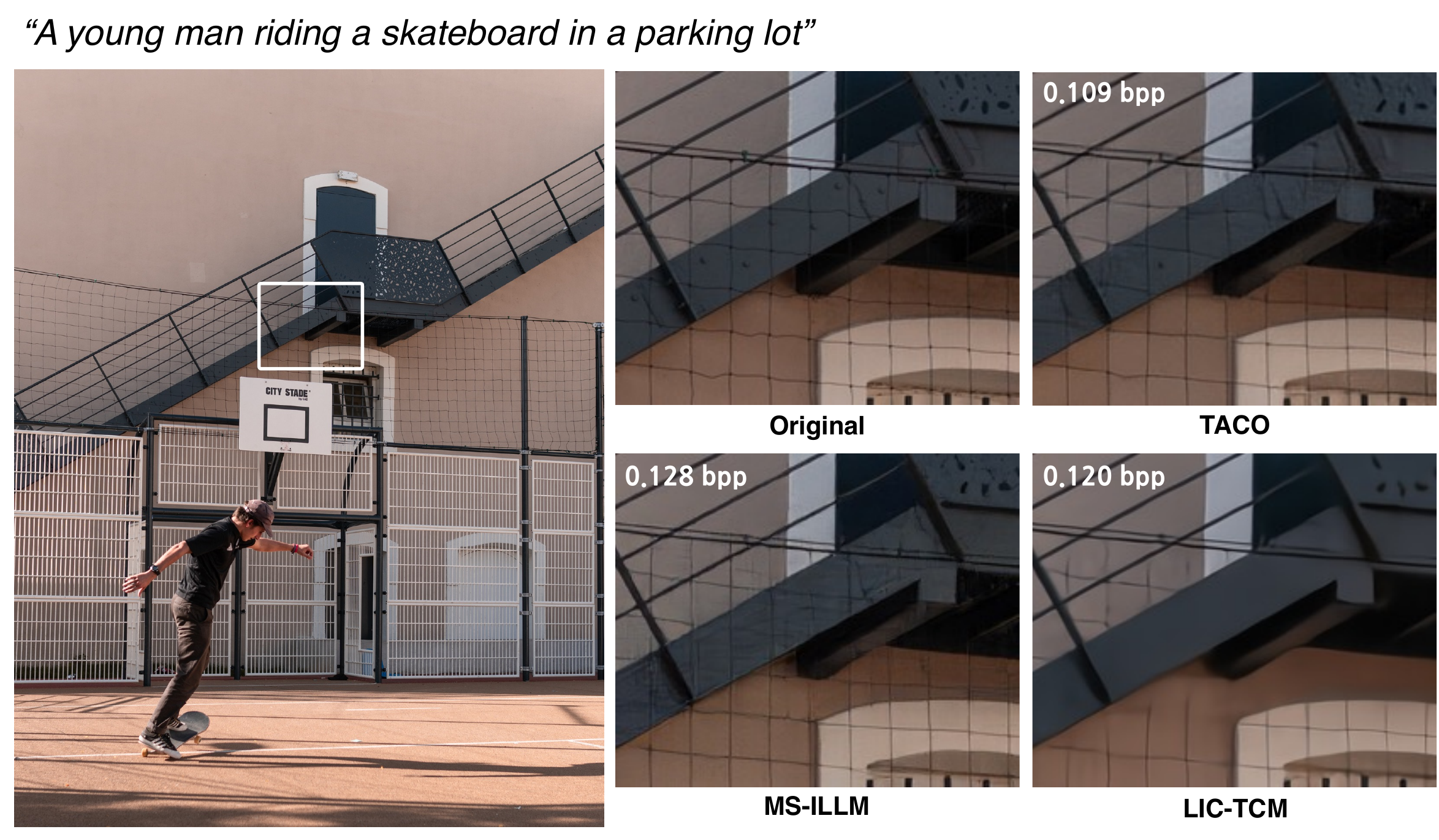}}\\ {\includegraphics[width=0.87\columnwidth]{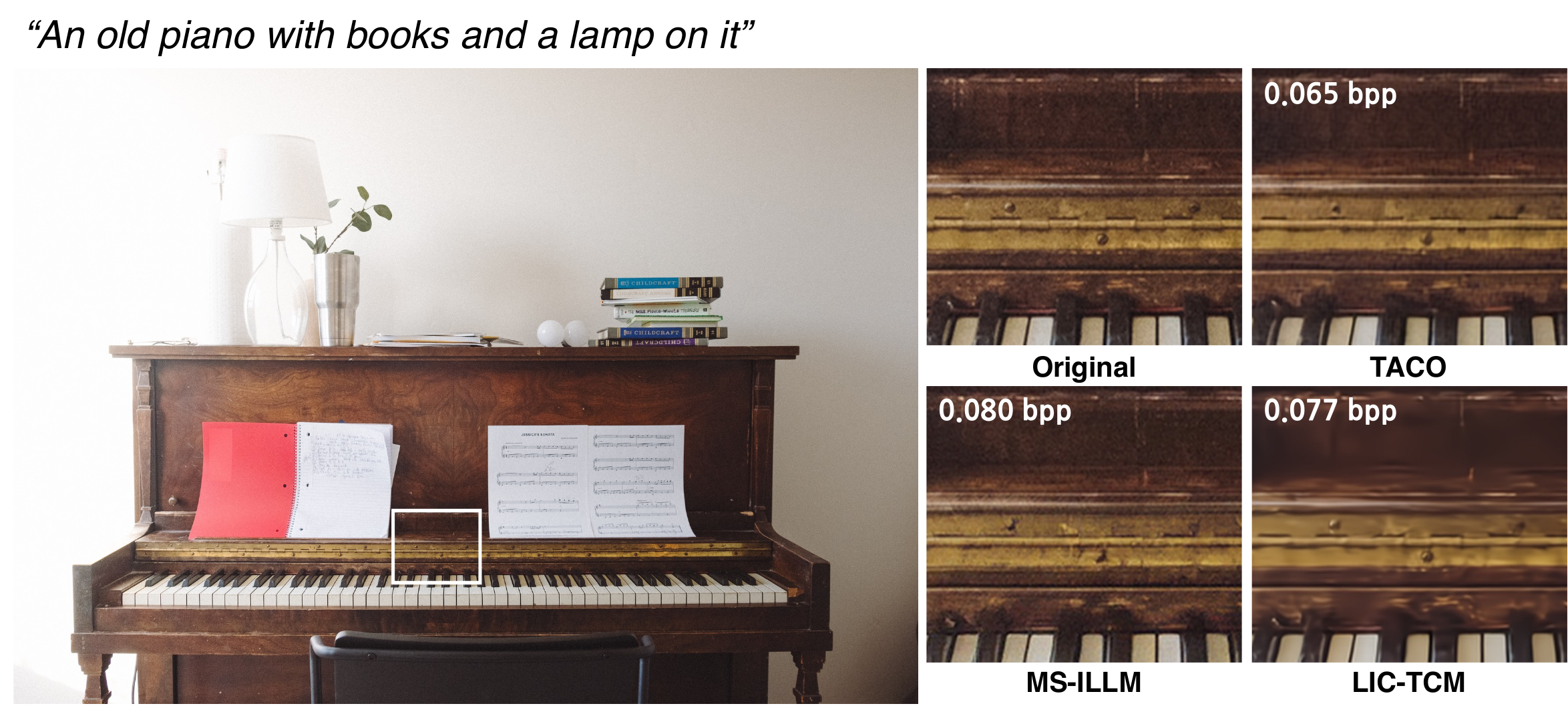}}\\ 
    \caption{\textbf{Additional Qualitative Results.}}

    \label{fig:qualitative_examples} 
\end{figure*}

\newpage
\begin{figure*}[!h]
    \centering 
    {\includegraphics[width=0.85\columnwidth]{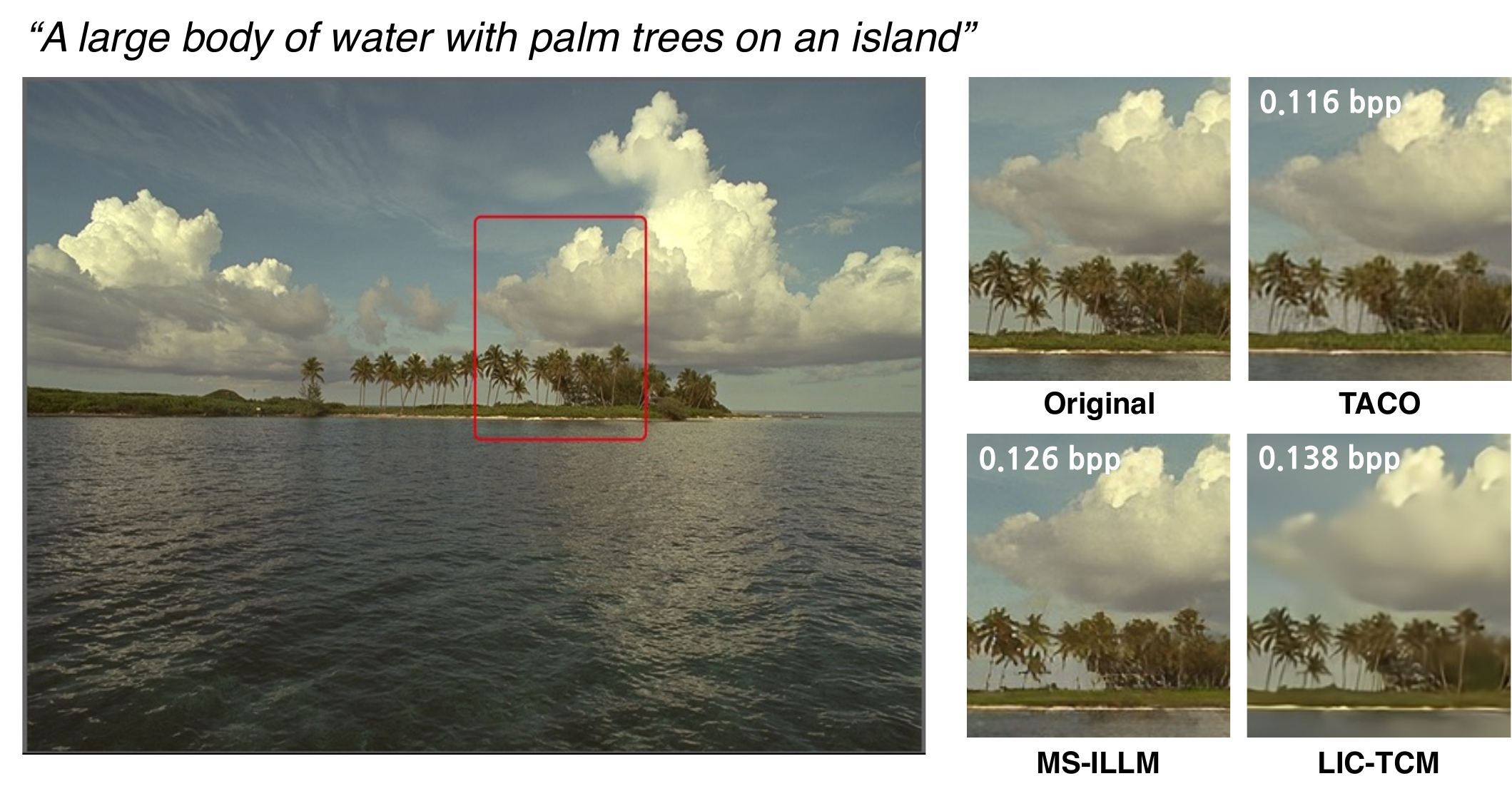}}\\      
    {\includegraphics[width=0.8\columnwidth]{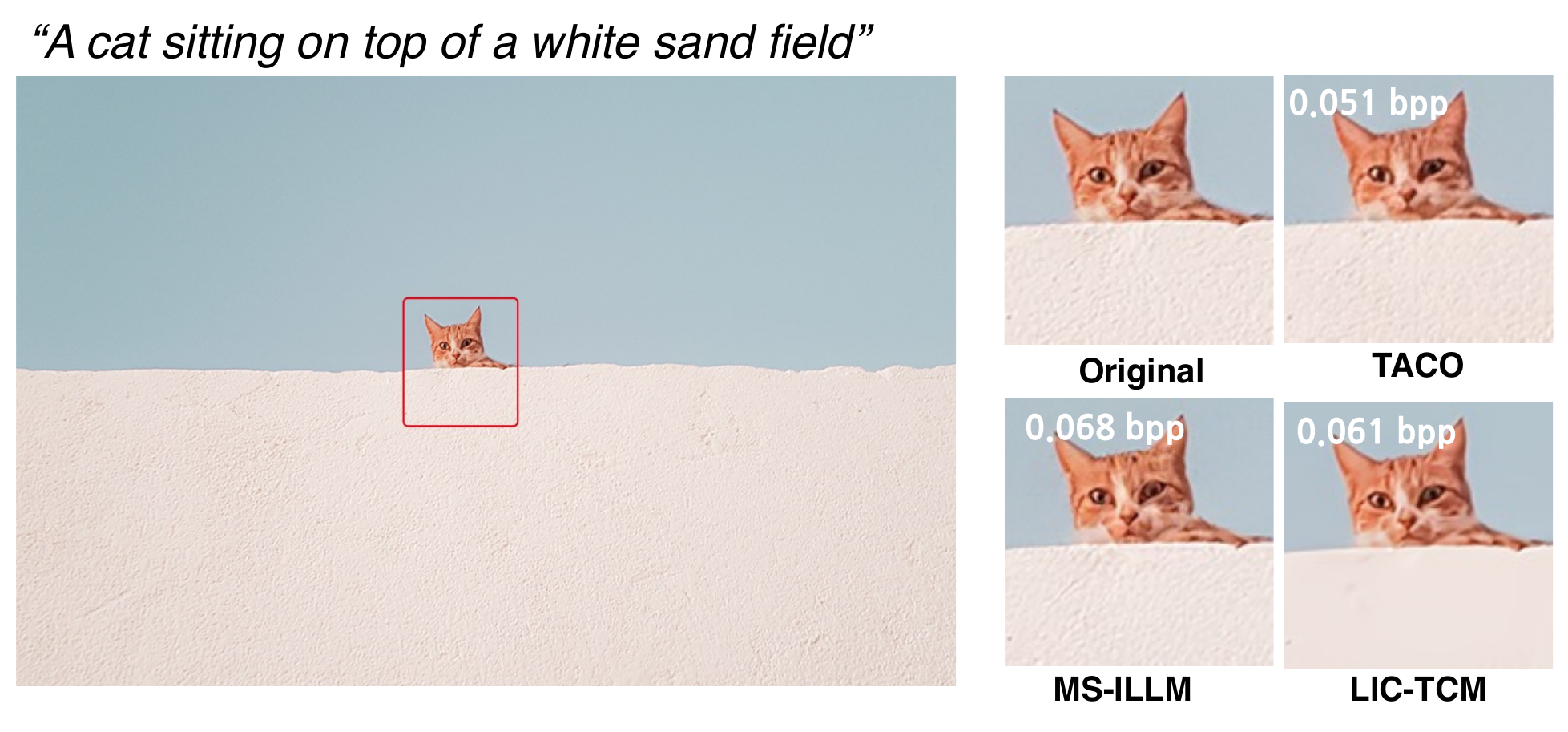}}\\ {\includegraphics[width=0.85\columnwidth]{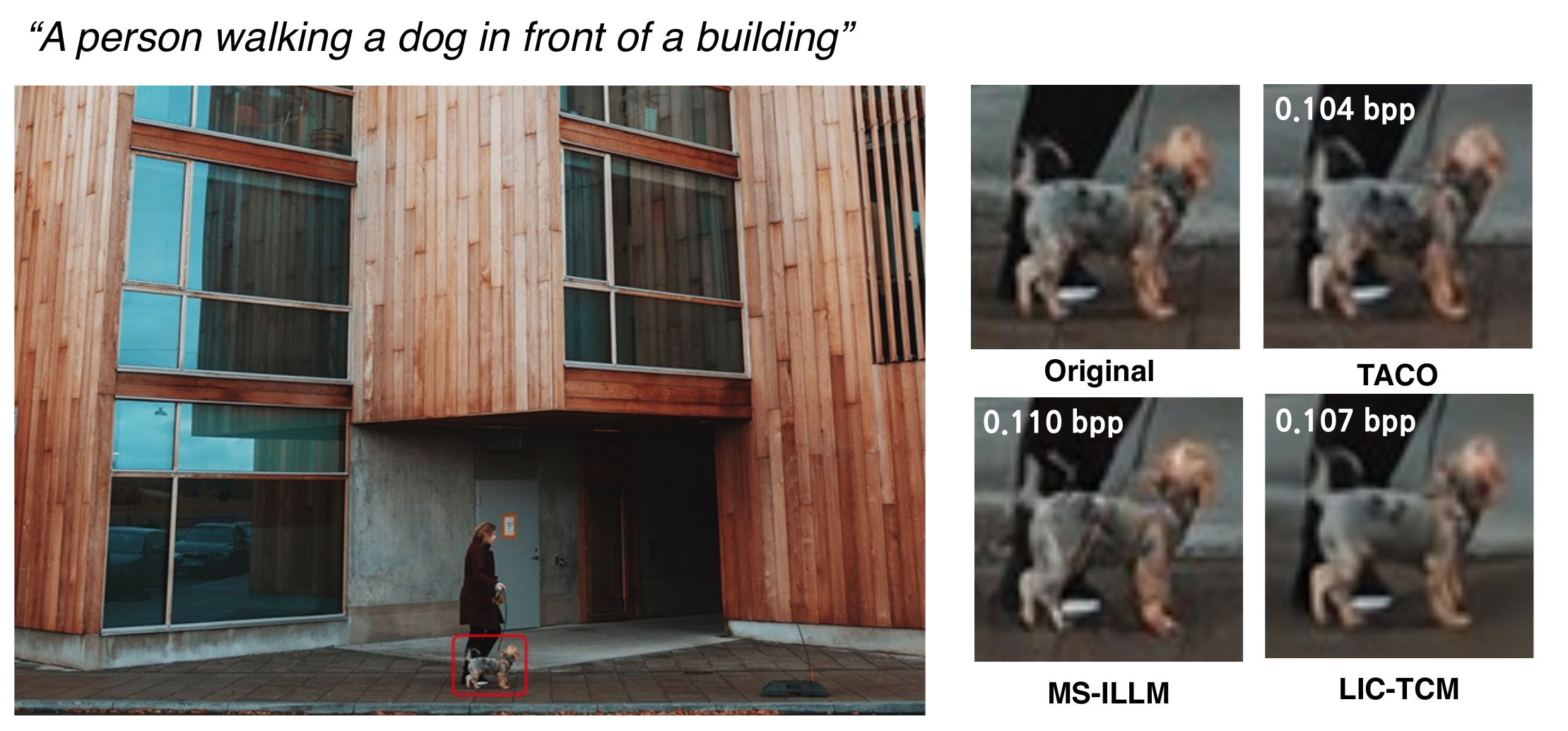}}\\ 
    \caption{\textbf{Additional Qualitative Results.}}

    \label{fig:qualitative_examples} 
\end{figure*}

%}

\end{document}